\documentclass{article}

\usepackage{PRIMEarxiv}
\usepackage[round]{natbib}

\usepackage{microtype}
\usepackage{graphicx}
\usepackage{booktabs}

\usepackage[hidelinks]{hyperref}

\usepackage{amsmath}
\usepackage{amssymb}
\usepackage{mathtools}
\usepackage{amsthm}

\usepackage[capitalize,noabbrev]{cleveref}

\usepackage{algorithm}
\usepackage{algorithmic}
\usepackage{nicefrac}
\usepackage{subcaption}
\usepackage{makecell}
\usepackage{collcell}
\usepackage{comment}

\DeclareMathOperator*{\argmin}{arg\,min} 
\newcommand{\ouracronym}{{\textsc{SFGE}}}    

\providecommand{\Description}[2][]{}

\usepackage{tikz}
\newcommand{\cblock}[3]{
  \hspace{-1.5mm}
  \begin{tikzpicture}[node/.style={square, minimum size=10mm, thick, line width=0pt}]
    \node[fill={rgb,255:red,#1;green,#2;blue,#3},scale=0.8] () [] {};
  \end{tikzpicture}%
}

\title{Score Function Gradient Estimation to Widen the Applicability of Decision-Focused Learning}

\author{
  Mattia Silvestri\thanks{Corresponding author. \texttt{mattia.silvestri94@gmail.com}} \\
  University of Bologna \\
  Bologna, Italy \\
  \And
  Senne Berden\thanks{Corresponding author. \texttt{senne.berden@kuleuven.be}} \\
  KU Leuven \\
  Leuven, Belgium \\
  \And
  Gaetano Signorelli \\
  University of Bologna \\
  Bologna, Italy \\
  \And
  Ali İrfan Mahmutoğulları \\
  KU Leuven \\
  Leuven, Belgium \\
  \And
  Jayanta Mandi \\
  KU Leuven \\
  Leuven, Belgium \\
  \And
  Brandon Amos \\
  Meta \\
  New York, USA \\
  \And
  Tias Guns \\
  KU Leuven \\
  Leuven, Belgium \\
  \And
  Michele Lombardi \\
  University of Bologna \\
  Bologna, Italy \\
}

\begin{document}
\maketitle

{\renewcommand{\thefootnote}{}%
\setlength{\footnotesep}{0pt}%
\footnotetext{\setlength{\parindent}{0pt}\noindent This article appears in the \emph{Journal of Artificial Intelligence Research}, Vol.~85 (2026). DOI: \href{https://doi.org/10.1613/jair.1.19498}{10.1613/jair.1.19498}. Available at \url{https://jair.org/index.php/jair/article/view/19498}. Please cite the published JAIR version.}%
\addtocounter{footnote}{-1}}

\begin{abstract}
{\bf Background:}
Real-world optimization problems often contain parameters that are unknown at solving time. For example, in delivery problems, these parameters may be travel times or customer demands.
A common strategy in such scenarios is to first predict the parameter values from contextual features using a machine learning model, and then solve the resulting optimization problem. To train the machine learning model, two paradigms can be distinguished. In \textit{prediction-focused learning}, the model is trained to maximize predictive accuracy. However, this can lead to suboptimal decision-making, because it does not account for how prediction errors affect the quality of the downstream decisions. 
To address this, \textit{decision-focused learning} (DFL) minimizes a task loss that captures how the predictions affect decision quality.

{\bf Objectives:}
One challenge in DFL is that the task loss has zero-valued gradients when the optimization problem is combinatorial, which hinders gradient-based training. For this reason, state-of-the-art DFL methods use surrogate losses and problem smoothing. However, these methods make specific assumptions about the problem structure (e.g., linear or convex problems with unknown parameters occurring only in the objective function).
The goal of our work is to overcome these limitations and extend the applicability of DFL.

{\bf Method:}
We propose an alternative DFL approach that makes only minimal assumptions by combining stochastic smoothing with score function gradient estimation.
This makes the approach broadly applicable, including to problems with nonlinear objectives, uncertainty in the constraints, and two-stage stochastic optimization problems.

{\bf Results:}
Our experiments show that our method matches or outperforms specialized methods for the problems they are designed for, while also extending to settings where no existing method is applicable.
In addition, our method always outperforms models trained with prediction-focused learning.

{\bf Conclusions:}
In this work we demonstrate that by combining stochastic smoothing and score function gradient estimation to estimate the gradients of a smoothed loss, we can train a machine learning model in a DFL fashion without assuming any structural property of the optimization problem.
This approach extends the applicability of DFL to a wider range of optimization problems, including those with uncertainty in the constraints. At the same time, it achieves performance that is competitive with or superior to existing DFL methods when they are applicable.

\end{abstract}

\section{Introduction}
\label{sec:intro}
Real-world optimization problems often contain parameters that are uncertain at solving time. Consider, for example, a manufacturing company that needs to schedule its production to meet uncertain customer demands, or a delivery company that needs to route its vehicles under uncertain traffic conditions. 
These kinds of problems can be framed as \textit{predict-then-optimize} problems, consisting of a prediction stage followed by an optimization stage.
In the prediction stage, a machine learning (ML) model is used to predict the unknown parameters from correlated contextual features. In the optimization stage, the optimization problem is solved using the predicted parameters. The quality of the resulting decisions is thus highly dependent on the ML model's predictions, making the procedure by which this model is trained particularly important. Two training paradigms can be distinguished: prediction-focused and decision-focused learning.

In \textit{prediction-focused learning} (PFL), the predictive model is trained to maximize the accuracy of the predicted parameters using traditional ML losses like the mean squared error or negative log-likelihood. One downside of this approach is that it can lead to suboptimal decision-making, because it does not account for the ways in which prediction errors can affect the solution to the optimization problem. The accuracy of the predictions does not necessarily align with the quality of the resulting decisions.
As an illustrative example, consider a knapsack problem in which the item values are unknown and must be predicted prior to solving. Overestimating the value of high-value items does not alter the optimal solution, since these items would be selected regardless of the overestimation. In contrast, underestimating their value by the same amount could cause the items to be excluded, leading to a different and potentially suboptimal solution.

For this reason, \textit{decision-focused learning} (DFL) instead trains the ML model to maximize the quality of the decisions resulting from the predictions by minimizing a task loss, and thus forms a deeper integration between prediction and optimization. Most DFL approaches are gradient-based, and therefore require backpropagation of the gradient through the optimization problem. While this can be done exactly 
for convex optimization problems using implicit differentiation \cite{amos2017optnet, agrawal2019differentiable}, combinatorial optimization problems present significant challenges.
This is because when the parameters of a combinatorial problem change, the solution either does not change at all, or changes abruptly.
Consequently, the partial derivatives of the solution with respect to the parameters are zero almost everywhere, and do not exist at the transition points. This makes the direct application of gradient-based ML techniques unhelpful.

To address this challenge, various techniques have been proposed to obtain non-zero gradients that still reflect decision quality \cite{donti2017task,elmachtoub2022smart,elmachtoub20a,MandiNEURIPS2020,mandi2022decision, mulamba2020contrastive,Shah0WPT22,aaai/WilderDT19, tang2024cave, berden2025solver}. However, most of these methods place strong assumptions on the structure of the optimization problem, for example, requiring that the predicted parameters appear only in the objective function (and  not in the constraints), that the problem's objective function is linear, or that there are no integrality constraints.



In this paper, we present a broadly applicable method that does not make such assumptions. Our method uses stochastic smoothing, by applying random perturbations to the predicted parameters at training time, according to a controllable distribution. 
This smooths out the loss and gives it informative gradients that can be used to train the predictive model in a decision-focused manner.
Moreover, by adjusting the distribution's parameters, the smoothed loss can asymptotically approach the true task loss, so that the location of any minimum can in principle be preserved.
Still, computing the gradients of the smoothed loss is not trivial without placing strong restrictions on the downstream problem. To overcome this, we propose to use score function gradient estimation (SFGE)~\cite{ReinforceWilliams92}.
This approach only requires the task loss to be bounded, allowing us to compute the gradient of the loss with respect to the parameters, regardless of whether they appear in the objective function, the constraints, or both. 
Furthermore, this approach can be used regardless of whether the problem is linear or not, and whether the problem contains integrality constraints, or other more involved constraints.

In our experimental evaluation, we start by showing that when the predicted parameters appear solely in the objective function, \ouracronym{} is marginally bested by the state of the art, while it still significantly outperforms prediction-focused methods. On the other hand, when the predictions (also) occur in the constraints, \ouracronym{} matches or outperforms the state of the art in terms of solution quality. 
We also demonstrate the effectiveness of our method on two-stage stochastic optimization problems, where it achieves great decision quality with vastly improved efficiency at inference time compared to the use of sample average approximation. Finally, we demonstrate that our method can be effectively applied to a nonlinear combinatorial optimization problem where no existing method can be used, further evidencing its broad applicability.

The remainder of this paper is organized as follows. In \Cref{sec:related_work}, we discuss related work. In \Cref{sec:prob_setting}, we formally define the predict-then-optimize problem setting, and the task losses used in the paper. We then introduce our method in \Cref{sec:method}, which we experimentally evaluate in \Cref{sec:exp_res}. Finally, \Cref{sec:conclusions} concludes the paper.


\section{Related Work}
\label{sec:related_work}
In this section we provide an overview of existing DFL methods.
We refer the reader to the survey papers by \citet{mandi2023decision} and \citet{sadana2025survey} for a more comprehensive discussion.

\paragraph{Differentiable Optimization}
Many DFL methods make use of differentiable optimization, which is concerned with computing the Jacobian of an optimization problem's optimal solution with respect to the problem's parameters. A pioneering work in differentiable optimization is OptNet by \citet{amos2017optnet}, which differentiates through convex quadratic programming problems using implicit differentiation of the Karush-Kuhn-Tucker (KKT) optimality conditions. Later, \citet{agrawal2019differentiating} and \citet[Chapter 7]{amos2019differentiable}
extended this approach by differentiating the optimality conditions of conic programs.
However, these methods are not directly applicable to combinatorial optimization problems, since the solution, and hence the task loss, is piecewise-constant in that setting, making the gradients zero almost everywhere.

\paragraph{Predicting objective parameters.}

Most existing works study the setting where the ML model only predicts parameters in the objective function of the optimization problem. \citet{aaai/WilderDT19} focus specifically on linear programs (LP) and make use of the OptNet method referred to above. However, because the optimal solution of an LP is a piecewise-constant function of its cost coefficients, differentiable optimization cannot be used straightforwardly. To this end, \citet{aaai/WilderDT19} analytically smooth the optimization problem by adding the Euclidean norm of the decision variables to the objective function, thereby resolving the zero-valued gradient issue. Similarly, \citet{MandiNEURIPS2020} employ log-barrier regularization for smoothing, and differentiate the homogeneous self-dual embedding of the LP, rather than the KKT conditions. For integer linear programs (ILPs) and mixed-integer linear programs (MILPs), both works relax the integrality constraints and thus make use of the LP relaxation of the problem. However, this relaxation can lead to suboptimal solutions in the DFL process, as we demonstrate Appendix~\ref{app:lp_vs_milp}. To improve on this, \citet{ferber2020mipaal} instead use a cutting planes approach that iteratively tightens the LP relaxation and leads to better decision quality as a result.

Another line of work achieves smoothing implicitly through perturbation, rather than by analytically altering the problem's formulation \citep{blackbox, berthet2020learning, niepert2021implicit}.
By perturbing the predicted objective parameters, they effectively smooth the mapping from the parameters to the solver’s output, leading to a non-zero Jacobian of the expected optimal solution. In \citep{blackbox}, the perturbation is deterministic and creates a linear interpolation of the piecewise-constant loss landscape, which produces informative non-zero gradients. In \cite{berthet2020learning}, the perturbations are sampled randomly, and the Jacobian is defined using the expected optimal solution over this noise distribution. These approaches enable the use of differentiable optimization layers, where the optimization problem itself is embedded as a component within a larger neural network architecture. 
However, unlike the approach we propose, they are not capable of handling problems in which the target parameters appear in the constraints, or nonlinearly in the objective function.

 A third line of work studies the use of surrogate losses that still reflect decision quality but offer informative non-zero gradients without requiring the use of differentiable optimization. A seminal paper is that of \citet{elmachtoub2022smart}, which proposed the SPO+ surrogate loss, a convex upper bound of the regret (a measure of decision quality that will be formalized in \Cref{sec:prob_setting}) that offers informative subgradients. Other notable examples include losses based on noise-contrastive estimation \citep{mulamba2020contrastive}, learning to rank \citep{mandi2022decision}, and geometric properties of the optimization problem's feasible region \citep{tang2024cave, berden2025solver}. Finally, the work by \citet{shah2022decision} proposes to \textit{learn} a surrogate loss from evaluations of the true task loss. To do so, they perform random perturbations of the true parameters in each training example, to fit convex local approximations of the decision loss around that example. These example-specific losses are then used during training. However, since the local losses are trained in isolation and are restricted to a specific functional form, this method may lead to an approximate loss function whose optima do not match those of the true task loss. This is further discussed in \Cref{appendix:lodl_counter}.

\paragraph{Predicting constraint parameters.}   
To the best of our knowledge, \textsc{CombOptNet} was the first work to enable the integration of an ILP problem with parameterized constraints as a layer of a neural architecture \cite{paulus2021comboptnet}. The resulting model can then be trained by minimizing any differentiable loss function. For example, the authors use the mean squared error (MSE) between the predicted and ground-truth solutions.
More recently, \citet{hu2023predict+} introduced a method to train a neural network to predict the cost and constraint parameters by minimizing the post-hoc regret, which is a measure of suboptimality computed after potentially applying a recourse action to ensure feasibility. The proposed approach relies on implicit differentiation and the interior point solver of \citet{MandiNEURIPS2020}. However, it can be applied only for linear packing and covering problems. 
In a follow-up work, 
\citet{hu2023twostage} proposed a method to predict parameters in the constraints for two-stage problems tackled via MILP. 
They consider the LP relaxations of the MILPs in both stages, which, as mentioned earlier, may lead to suboptimal solutions.
\citet{hu2023branch} later proposed a method applicable to every iteratively solvable problem. Nevertheless, this method assumes a linear predictive model and employs coordinate descent for training, and thus cannot be applied when the predictive model is a more complex neural network. Finally, a recent paper by \citet{mandi2025feasibilityawaredecisionfocusedlearningpredicting} takes a different approach than minimizing the post-hoc regret, by avoiding explicit recourse modeling and instead focusing on reducing infeasibility. They propose two loss functions: one penalizes predicted solutions that are infeasible, while the other penalizes true solutions that become infeasible under the predicted parameters. By tuning the relative weights of these losses, they achieve mostly feasible solutions with small optimality gaps. As this work appeared after our initial submission, we did not compare against it.




\section{Problem Setting}
\label{sec:prob_setting}
We consider a parametric optimization problem:
%
\begin{equation}
\label{eq:opt_generic}
z^\star(y) = \argmin_{z \in Z(y)} f(z, y) 
\end{equation}
The optimization problem returns an optimal solution $z^\star(y) \in \mathbb{R}^n$, which is a minimizer of the objective function $f(.,y)$ within the feasible set $Z(y)$.
We make no assumption on $Z(y)$, which can include integrality constraints.
The optimization problem can be viewed as a mapping $y \mapsto  z^\star(y)$ because $z^\star(y)$ depends on $y$ through $f(.,y)$  and $Z(y)$. 
In predict-then-optimize problems, the true parameter vector $y$ is unknown, but correlated to observable features $x$.
We view both $x$ and $y$ as realizations for two random variables $X$ and $Y$ with a joint distribution $P(X, Y)$.
A dataset of samples $\{x_i, y_i\}_{i=1}^N$ drawn from $P(X, Y)$ is assumed to be available, and is used to train an ML model $m_\omega$ to predict $y$ based on $x$. At inference time, the model is invoked to obtain predictions $\hat{y} = m_\omega(x)$, which are then used to obtain a solution vector $z^\star(\hat{y})$.

\Cref{eq:opt_generic} differs from most predict-then-optimize setups in that no assumption is made on the cost function, and that the parameters are also allowed to appear in the constraints. Note that
this also encompasses settings where $y$ appears only in $f(.,y)$ or only in $Z(y)$.

\paragraph{Task losses.}

When the unknown parameters $y$ only occur in the \textit{objective} (i.e., $Z(y)$ is a fixed set $Z$), the quality of a solution $z^\star(\hat{y})$ can be evaluated in terms of \textit{regret}, also referred to as the smart predict-then-optimize (SPO) loss by \citet{elmachtoub2022smart}, which expresses the suboptimality of $z^\star(\hat{y})$ (the decisions made on the basis of the predicted parameters $\hat{y}$) with respect to the true parameters $y$:
\begin{equation}
  \label{eq:regret}
    \mathit{Regret}(\hat{y}, y) = f(z^\star(\hat{y}), y) - f(z^\star(y), y) 
\end{equation}
When the unknown parameters occur in the \emph{constraints}, the solution $z^\star(\hat{y})$ may violate the true constraints, i.e.,  $z^\star(\hat{y}) \notin Z(y)$. 
For example, production volumes in a manufacturing context might fail to meet demands when those are uncertain.
In practical cases, constraint violation must be corrected to recover feasibility, thus changing an infeasible solution $z^\star(\hat{y})$ into a feasible solution $z^\star_{corr}(\hat{y}, y)$. For example, in the manufacturing context mentioned above, this may involve buying additional products from a more expensive source in order to meet demand.
In the stochastic optimization literature, this is known as a \emph{recourse action} \citep{birge2011}. Recourse actions often have an associated cost, which can be formally taken into account by introducing a (problem-specific) penalty function $\mathit{Pen}(z^\star(\hat{y}), z^\star_{corr}(\hat{y}, y))$ that expresses the cost of correcting an infeasible solution $z^\star(\hat{y})$ to a feasible solution $z^\star_{corr}(\hat{y}, y)$.
With these concepts in place, we are ready to define the \textit{post-hoc regret}, introduced by \citet{hu2023branch, hu2023predict+, hu2023twostage}:
%
\begin{align}
  \label{eq:pregret}
    \mathit{PRegret}(\hat{y}, y) = f(z^\star_{corr}(\hat{y}, y), y) - f(z^\star(y), y) + \mathit{Pen}(z^\star(\hat{y}),  z^\star_{corr}(\hat{y}, y))
\end{align}
%
Note that, in what follows, we will not always distinguish between the regret and the post-hoc regret. We will generally only refer to the post-hoc regret, without loss of generality. This can be done because when the solution $z^\star(\hat{y})$ is feasible (which is always the case when $\hat{y}$ occurs only in the objective), the regret and post-hoc regret are identical.

\paragraph{Training objective.}

The goal is to train ML model $m_\omega$ to minimize the \emph{expected} post-hoc regret, leading to the following formulation:
\begin{equation}
    \label{eq:training}
    \argmin_\omega \mathbb{E}_{x, y \sim P(X, Y)} \left[ \mathit{PRegret}(m_\omega(x), y) \right]
\end{equation}

\section{Score Function Gradient Estimation}
\label{sec:method}
The challenge in minimizing the expected post-hoc regret \eqref{eq:training} through gradient-based training, is that for many problem classes of interest, the gradients of the post-hoc regret are zero-valued. This can be seen by applying the chain rule:
%
%
\begin{equation}
    \frac{\partial \mathit{PRegret}(\hat{y}, y)}{\partial \omega} = \frac{\partial\mathit{PRegret}(\hat{y}, y)}{\partial \hat{y}} \frac{\partial \hat{y}}{\partial \omega}
    \label{eq:chain_rule}
\end{equation}
The second factor $\nicefrac{\partial \hat{y}}{\partial \omega}$ relates only to the output of the predictive model and forms no issue. The first factor is more complicated:
\begin{subequations}
\begin{align}
    \frac{\partial  \mathit{PRegret}(\hat{y}, y) }{\partial \hat{y}} =\; &\frac{\partial(  f(z^\star_{corr}(\hat{y}, y), y) - f(z^\star(y), y) + \mathit{Pen}(z^\star(\hat{y}), z^\star_{corr}(\hat{y}, y))}{\partial \hat{y}} \\
    =\; &\frac{\partial  f(z^\star_{corr}(\hat{y}, y), y)}{\partial{\hat{y}}} + \frac{\partial\mathit{Pen}(z^\star(\hat{y}), z^\star_{corr}(\hat{y}, y))}{\partial \hat{y}} \\
    =\; &\frac{\partial  f(z^\star_{corr}(\hat{y}, y), y)}{\partial{z^\star_{corr}(\hat{y}, y)}} \frac{\partial z^\star_{corr}(\hat{y}, y)}{\partial z^\star(\hat{y})}\frac{\partial z^\star(\hat{y})}{\partial \hat{y}} +
    \frac{\partial\mathit{Pen}(z^\star(\hat{y}), z^\star_{corr}(\hat{y}, y))}{\partial z^\star(\hat{y})} \frac{\partial z^\star(\hat{y})}{\partial \hat{y}} +  \label{eq:chain_ruled_gradient} \\&\frac{\partial\mathit{Pen}(z^\star(\hat{y}), z^\star_{corr}(\hat{y}, y))}{\partial z_{corr}^\star(\hat{y}, y)} \frac{\partial z^\star_{corr}(\hat{y}, y)}{\partial z^\star(\hat{y})}\frac{\partial z^\star(\hat{y})}{\partial \hat{y}} \nonumber
\end{align}
\end{subequations}
The factor $\frac{\partial z^\star (\hat{y})}{ \partial \hat{y}}$ occurs in every term of \Cref{eq:chain_ruled_gradient}, and measures the change in $z^\star (\hat{y})$ when $\hat{y}$ changes infinitesimally. If the problem is combinatorial, this change is zero almost everywhere since small changes in problem parameters do not affect the optimal solution.
This makes the gradient $\frac{\partial \mathit{PRegret}(\hat{y}, y)}{\partial \omega}$ zero almost everywhere, and thus obstructs gradient-based learning.


\paragraph{Stochastic smoothing.}
To tackle this issue, we apply a stochastic perturbation to the predictor output during training.
Formally, at training time the ML model no longer outputs a point estimate $\hat{y}$, but a pair $\theta = (\mu, \sigma)$ defining a Gaussian distribution $p_\theta(\hat{y})$ over the problem parameters, centered on the original point estimate and having a standard deviation equal to $\sigma$.
This leads to a new, smoothed loss function, as an expectation over $p_\theta(\hat{y})$:
\begin{equation}
\label{eq:dist-loss}
L(\theta, y) = \mathbb{E}_{\hat{y} \sim p_{\theta}(\hat{y})} [\mathcal{L}(\hat{y}, y)]
\end{equation}
where $\mathcal{L}$ refers to the original task loss, i.e., the (post-hoc) regret.

\begin{figure}[t]
  \centering
  \includegraphics[width=0.5\linewidth]{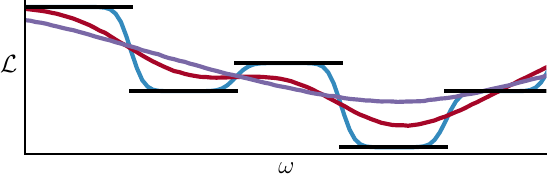}
  \caption{Illustration of a DFL loss with non-informative zero-valued derivatives (\protect\cblock{0}{0}{0}) smoothed by a Gaussian distribution over the parameters, with increasing variances (\protect\cblock{52}{138}{189} $\leq$ \protect\cblock{166}{6}{40} $\leq$ \protect\cblock{128}{114}{179}). The larger the variance, the more the loss gets smoothed, but the less it resembles the original piecewise-constant task loss.}
  \label{fig:dists}
  \Description{Three lines that show the effect of smoothing the regret.}
\end{figure}

The smoothness of $L(\theta, y)$ is controlled by $\sigma$, as depicted in \Cref{fig:dists}.
As $\sigma$ decreases, the expectation asymptotically converges to the true task loss.
As a consequence, the location of every optimum can be preserved, at the cost of less informative gradients. As $\sigma$ increases, the smoothed loss increasingly diverges from the true task loss. Intermediate values offer a trade-off.
It is also possible to treat $\sigma$ as a learnable parameter, which we do in our experiments, and discuss in more detail in Appendix~\ref{appendix:distr_and_grad}. With this setup, the training process is left free to adjust the degree of smoothing so as to minimize the expected loss. In Appendix~\ref{appendix:sigma_behavior}, we provide more insight into the learned behavior of $\sigma$, and show that it tends to decrease throughout training, which can be interpreted as a shift from exploration to exploitation.
Note that even though we use a Gaussian distribution for smoothing, we do not actually assume the data to be normally distributed, nor do we expect to accurately learn the variance in the data: the purpose of $p_\theta(\hat{y})$ is purely that of smoothing the task loss.

\paragraph{Training with gradient estimation.}

The smoothed loss function $L(\theta, y)$ now has an informative non-zero gradient $\frac{\partial L(\theta, y)}{\partial \theta}$. However, the computation of this gradient is not trivial. To compute an estimate, we utilize score function gradient estimation (SFGE), also known as the REINFORCE algorithm in the context of reinforcement learning \cite{Monte_Carlo_gradient_estimation}.
Consider the following derivation:
\begin{subequations}
\begin{align}
\nabla_{\theta}L(\theta, y) &= \nabla_{\theta}\mathbb{E}_{\hat{y} \sim p_{\theta}(\hat{y})} [\mathcal{L}(\hat{y}, y)] \label{eq:integral}\\
&= \nabla_{\theta}\int p_{\theta}(\hat{y}) \mathcal{L}(\hat{y}, y) d\hat{y} \label{eq:expectation_definition}\\
&= \int \mathcal{L}(\hat{y}, y) \nabla_{\theta}p_{\theta}(\hat{y}) d\hat{y}\label{eq:bring_gradient_inward}\\
&= \int \mathcal{L}(\hat{y}, y) p_{\theta}(\hat{y}) \nabla_{\theta}\log p_{\theta}(\hat{y}) d\hat{y} \label{eq:log_derivative}\\
&= \mathbb{E}_{\hat{y} \sim p_{\theta}(\hat{y})} [\mathcal{L}(\hat{y}, y) \nabla_{\theta}\log p_{\theta}(\hat{y})] \label{eq:result_derivation}
\end{align}
\end{subequations}
where \eqref{eq:integral} follows from the definition of the loss in \eqref{eq:dist-loss}, \eqref{eq:expectation_definition} and \eqref{eq:result_derivation} follow from the definition of the expectation and \eqref{eq:log_derivative} follows the log derivative trick, i.e., the identity $\nabla_{\theta}\log p_{\theta}(\hat{y}) = \frac{\nabla_{\theta}p_{\theta}(\hat{y})}{p_{\theta}(\hat{y})}$ whenever $p_{\theta}(\hat{y}) > 0$. The final gradient in \eqref{eq:result_derivation} can be estimated using a Monte Carlo method, giving:
\begin{align}
\begin{split}
\label{eq:monte_carlo_approximation}
    \nabla_{\theta} L(\theta, y) &\approx \frac{1}{S} \sum_{i=1}^{S} \mathcal{L}(\hat{y}^{(i)}, y) \nabla_{\theta} \log p_{\theta}(\hat{y}^{(i)}) \\
\end{split}
\end{align}
with $\hat{y}^{(i)} \sim p_{\theta}(\hat{y})$ and $S$ is the total number of samples.
We empirically observed that a single sample (i.e., $S = 1$) is typically sufficient, as the gradient noise is mitigated by using stochastic gradient descent. We expand on this in Appendix~\ref{appendix:distr_and_grad}. We also note that SFGE is known to suffer from high variance \cite{greensmith2004variance}, which can slow down the speed of convergence. To mitigate this issue, we employ \emph{standardization of the regret on each mini-batch}, as also detailed in \Cref{appendix:distr_and_grad}.

The validity of the interchange of the integral and gradient in \eqref{eq:bring_gradient_inward} requires deeper analysis. As discussed in \citet{Monte_Carlo_gradient_estimation}, the interchange is valid if:
\begin{itemize}
    \item[(i)] $p_{\theta}(y)$ is continuously differentiable in its parameters $\theta$, 
    \item[(ii)] $p_{\theta}(\hat{y}) \mathcal{L}(\hat{y}, y)$ is both integrable and differentiable for all parameters $\theta$, and 
    \item[(iii)] There exists an integrable function $g(\hat{y})$ such that $\sup_{\theta}\| \mathcal{L}(\hat{y}, y)  \nabla_{\theta} p_{\theta}(\hat{y})\|_1 \leq g(\hat{y}), \forall \hat{y}$. 
\end{itemize}
For an arbitrary choice of probability density $p_\theta$ and task loss $\mathcal{L}$, it is difficult to check that these three conditions hold (see \citet{l1995note} or  \citet{glasserman1990gradient} for a more detailed discussion). Therefore, we make nonrestrictive assumptions to prove that the above conditions hold for our case. 

First, since $p_\theta$ is normally distributed, the condition (i) holds due to the smoothness of the density function. For the univariate case:
\begin{equation*}
     p_{\theta}(y) = \frac{1}{\sigma\sqrt{2\pi}}e^{-{\frac {1}{2}}\left({\frac {y-\mu }{\sigma }}\right)^{2}}
\end{equation*}
where $\theta = (\mu, \sigma)$. Then, the derivatives of the normal distribution with respect to $\mu$ and $\sigma$ are
\begin{equation}\label{eq:univariate_gradients}
\frac{y - \mu}{\sigma^2} p_{\theta}(y) \quad \text{and}
\quad \left (\frac{(y-\mu)^2}{\sigma^3} - \frac{1}{\sigma}\right)p_{\theta}(y), 
\end{equation}
respectively. 
Both derivatives are continuous in their respective parameters, unless $\sigma=0$, which can easily be avoided by either using a fixed smoothing parameter $\sigma$, or by adding a small positive constant to a trainable, non-negative, $\sigma$. The extension to the multivariate case follows directly. Thus (i) holds.

In order to show that (ii) holds, we can also assume that the post-hoc regret $\mathit{PRegret}(\hat{y}, y)$ is bounded. The first part of the post-hoc regret is bounded as it is the difference between the objective values of two feasible solutions if the feasible region of the optimization problem is also bounded. We can also assume that the second part, the penalty, always gives a finite value and admits an upper bound, since in practice there is no infinitely infeasible decision to correct. Then, since $\mathcal{L}$ is bounded, and under the Gaussian assumption, the product $p_{\theta}$ and $\mathcal{L}$ is integrable and differentiable for all parameters $\theta$. Hence (ii) holds under these assumptions. 

To show (iii), first note that $\nabla_{\theta} p_{\theta}(\hat{y})$ takes a finite value for all $\theta$, which covers our case since $\mu$ matches the output of the original point estimator model and $\sigma$ is controllable.
The univariate case is clear in \eqref{eq:univariate_gradients} and the extension to the multivariate case is similar. Therefore, $\nabla_{\theta} p_{\theta}(\hat{y})$ is bounded, i.e., there exists a (possibly large) positive real number $M(\hat{y})$  depending on $\hat{y}$ such that $\sup_{\theta}\| \nabla_{\theta} p_{\theta}(\hat{y})\|_1 \leq M(\hat{y})$ for all $\hat{y}$.

Using the Cauchy–Schwarz inequality, we get:
\begin{equation*}
    \|\mathcal{L}(\hat{y}, y)  \nabla_{\theta} p_{\theta}(\hat{y})\|_1 \leq  \mathcal{L}(\hat{y}, y)  \| \nabla_{\theta} p_{\theta}(\hat{y})\|_1
\end{equation*}
for all $\theta$ and $\hat{y}$ since $\mathcal{L}$ is a non-negative real-valued function. Taking the supremum of both sides of the equation  with respect to $\theta$, we get:
\begin{equation*}
    \sup_{\theta} \| \mathcal{L}(\hat{y}, y)  \nabla_{\theta} p_{\theta}(\hat{y})\|_1 \leq  \mathcal{L}(\hat{y}, y)   \sup_{\theta}\| \nabla_{\theta} p_{\theta}(\hat{y})\|_1 
\end{equation*}
since the loss does not depend on $\theta$.  Then, we have:
\begin{equation*}
    \sup_{\theta}\| \mathcal{L}(\hat{y}, y)  \nabla_{\theta} p_{\theta}(\hat{y})\|_1 \leq g(\hat{y}) := \mathcal{L}(\hat{y}, y)  M(\hat{y})
\end{equation*}
where $g$ is constant and hence integrable. Hence (iii) holds. 

Our approach differs from existing DFL methods in its broad applicability, due to the fact that \Cref{eq:monte_carlo_approximation} makes no assumptions about the optimization problem form or the location of the predicted parameters. 
In our experimentation, we consider settings \emph{with linear and nonlinear objectives, with and without integrality constraints, and involving uncertain parameters appearing in the objective, in the constraints, or in both}.

\paragraph{Test time.}
The prediction of a \textit{distribution} $p_\theta(\hat{y})$ through the prediction of $\mu$ and $\sigma$ is introduced solely for the purpose of obtaining informative non-zero gradients. At test time, we directly feed point predictions $\mathbb{E}_{\hat{y} \sim p_\theta}[\hat{y}] = \mu$ from the ML model to the optimization problem. This is in contrast with classical stochastic optimization approaches, which would learn a distribution over parameter vectors, from which multiple realizations would be sampled and incorporated in one large deterministic optimization problem, a process known as sample average approximation (SAA). Our approach thus offers large scalability improvements at inference time with respect to SAA, which we will evidence in our experimental evaluation.

Two notes are in order. The first is that, because of the use of point predictions at test time, our method may not be appropriate for problems whose optimal solutions cannot be identified via a single predicted vector, as studied in \citet{schutte2025sufficient}. In such cases, our formulation will be structurally suboptimal, while scenario-based methods can achieve asymptotic optimality, given unlimited data and computation time. The second is that, this use of point predictions at test time justifies the use of a Gaussian distribution for smoothing, even when the true data distribution is known to be non-Gaussian. After all, only the mean of the predicted distribution is used at test time. Thus, accurately fitting the remainder of the distribution offers no additional benefits, unless a scenario-based method is employed instead.

\section{Experimental Results}
\label{sec:exp_res}
To demonstrate the generality of our approach, we conducted the experimental analysis by focusing on four research questions:
\begin{enumerate}
    \item How does \ouracronym{} compare with PFL and state-of-the-art DFL approaches when predicting parameters appearing exclusively in the constraints, or in both the objective and constraints?
    \item How does \ouracronym{} fare against a PFL method that solves a two-stage stochastic optimization approach via SAA at inference time?
    \item How does \ouracronym{} compare with PFL and state-of-the-art DFL approaches when predicting parameters exclusively in the objective function?
    \item How does \ouracronym{} perform on a problem that falls outside the scope of existing DFL methods?
\end{enumerate}

In our experimental evaluation, we use linear regression models for each method. This is common practice in DFL evaluations and is done to obtain a misspecified predictive model -- a setting in which DFL is most promising~\cite{elmachtoub2022smart, mandi2020smart, mandi2022decision, mandi2023decision, schutte2023robust, berden2025solver}.
When utilizing \ouracronym{}, the model predicts the mean of a Gaussian distribution, from which we draw one sample of $\hat{y}$ per gradient estimation (i.e., $S=1$), which we found to work best in practice (see Appendix~\ref{appendix:distr_and_grad}).
For each dataset, we use a training-validation-test split of 80\%, 10\%, and 10\%, respectively.
We refer the reader to the appendices for more details about implementation and related discussions. 
All the experiments were run on a machine equipped with an Intel(R) Core(TM) i7-1065G7 1.30GHz CPU and 16GB of RAM. All code and data are available at \url{https://github.com/matsilv/sfge-dfl}.

\subsection{Q1: Predicting Constraint Parameters}

We start with the task of predicting parameters in the constraints, a challenging problem that has received limited attention in existing work on DFL.

\paragraph{Linear programs.}
The approach by \citet{hu2023predict+} specifically focuses on packing and covering LPs.
Thus, in our first experiment, we consider a packing LP -- a fractional knapsack problem (KP) with 10 items. In this KP benchmark, both the item values and the item weights are unknown and must be predicted from correlated features. The features are synthetically generated using the procedure described by \citet{hu2023predict+}. This benchmark allows us to compare \ouracronym{} against a method that is best-in-class for this specific setting, which we refer to as IntOpt-C in the tables of results.

We use the same recourse action and penalty functions as in \citet{hu2023predict+}: when the solution instantiated by the prediction exceeds the capacity, the amount of each item selected is scaled down until the capacity constraint is satisfied. When the discarded amount of item $i$ is $\Delta_i$, the associated penalty for removing it is $\rho v_i \Delta_i$, where $v_i$ is the item's value.
We consider problems with a capacity of 50 and penalty coefficients $\rho = \{ 0, 1, 2 \}$.
We conduct experiments on 10 different training-validation-test splits.
All methods are trained with Adam~\cite{kingma2014adam}, a learning rate of 0.005 and a batch size of 32 samples. Training is stopped when the validation regret (for DFL methods including \ouracronym{}) or the validation MSE (for PFL) has stopped improving.

\begin{table*}
\centering
\caption{Q1: Predicting Constraints Parameters - Linear programs. PFL, \textsc{\ouracronym{}} and \textsc{PO} results on the fractional KP. We omit the \textit{Feas. rel. PRegret} for \textit{Infeas. ratio}s near 1.}
\label{table:frational_kp}
\addtolength{\tabcolsep}{2pt}
\small  
    \begin{tabular}{lccccc}
    
    \toprule
    
    \textit{Method}    & \textit{Rel. PRegret}        & \textit{Feas. rel. PRegret}          & \textit{Infeas. ratio}       & \textit{MSE}                                                    & \textit{Epochs} \\

    \midrule
    
    \multicolumn{6}{c}{capacity=50, $\rho=0$} \\

    \midrule
        
        \textsc{PFL}  & $ 0.403 \pm 0.015 $           & $ \mathbf{0.107 \pm 0.064} $         & $ \mathbf{0.72 \pm 0.15} $   & $ \mathbf{99.1 \pm 13.1} $                                          & $ 13.3 \pm 2.9 $ \\
        IntOpt-C  & $\mathbf{0.377 \pm 0.130} $            & $ - $                                & $ 1.00 \pm 0.0 $             & $ 9.8 \cdot 10^5 \pm 1.2 \cdot 10^4 $                               & $ \mathbf{2.5 \pm 1.9} $      \\
        \textsc{SFGE} (ours) & $ 0.385 \pm 0.008 $    & $ - $                                & $ 1.00 \pm 0.0 $             & $ 8.2 \cdot 10^5 \pm 6.8 \cdot 10^5 $                               & $ 13.4 \pm 3.7 $ \\
    
    \midrule
    
    \multicolumn{6}{c}{capacity=50, $\rho=1$} \\
    
    \midrule
        
        \textsc{PFL}  & $ 0.501 \pm 0.033 $           & $ \mathbf{0.107 \pm 0.064} $         & $ 0.72 \pm 0.15 $            & $ \mathbf{99.1 \pm 13.1} $                                          & $ 13.3 \pm 2.9 $ \\
        
        IntOpt-C  & $ \mathbf{0.460 \pm 0.162} $           & $ 0.380 \pm 0.018 $                  & $ 0.61 \pm 0.02 $            & $ 3.8 \cdot 10^5 \pm 4.8 \cdot 10^3 $                               & $ \mathbf{2.1 \pm 1.9} $ \\
        \textsc{SFGE} (ours) & $ 0.467 \pm 0.016 $    & $ 0.177 \pm 0.045 $                  & $ \mathbf{0.55 \pm 0.10} $   & $ 7.9 \cdot 10^5 \pm 5.1 \cdot 10^5 $                               & $ 14.9 \pm 4.7 $ \\

    \midrule
    
    \multicolumn{6}{c}{capacity=50, $\rho=2$} \\
    
    \midrule
        
        \textsc{PFL}  & $ 0.600 \pm 0.077 $            & $ \mathbf{0.107 \pm 0.064} $         & $ 0.72 \pm 0.15 $           & $ \mathbf{99.1 \pm 13.1} $                                          & $ 13.3 \pm 2.9 $ \\
        IntOpt-C  & $\mathbf{0.492 \pm 0.173}$              & $ 0.422 \pm 0.009 $                  & $ \mathbf{0.42 \pm 0.05} $  & $ 3.5 \cdot 10^5 \pm 3.8 \cdot 10^3 $                               & $ \mathbf{1.6 \pm 0.6} $ \\
        \textsc{SFGE} (ours) & $ 0.512 \pm 0.036 $     & $ 0.237 \pm 0.092 $                  & $ 0.46 \pm 0.18 $           & $ 1.3 \cdot 10^6 \pm 9.3 \cdot 10^5 $              
                      & $ 16.8 \pm 4.3 $ \\

    \bottomrule
    
    \end{tabular}
\end{table*} 

The results are presented in \Cref{table:frational_kp} (additional results are given in \Cref{appendix:supp_res}). For each method, we report the relative post-hoc regret (\textit{Rel. PRegret}), the relative regret of solutions not requiring recourse actions (\textit{Feas. rel. regret}), the ratio of solutions that require recourse actions (\textit{Infeas. ratio}), the MSE, and the number of epochs until convergence. 
Both DFL methods outperform PFL in terms of post-hoc regret. Although IntOpt-C has lower post-hoc regret than SFGE, it exhibits notably higher variance. Our method performs slightly worse on average, but with much lower variance.
IntOpt-C is fastest in terms of convergence speed, whereas \textsc{PFL} and \ouracronym{} are slower and require a comparable number of epochs. As expected, PFL delivers the best MSE, because it is trained for maximal accuracy. 
 With increasing $\rho$, the DFL methods become more conservative: the infeasibility ratio decreases, but at the cost of a worse relative regret on the feasible solutions.

\paragraph{Integer linear programs.}
While IntOpt-C is limited to linear packing and covering problems, 
it has been extended for more general LPs by \citet{hu2023twostage}. 
However, many real-world combinatorial optimization problems entail integrality constraints and can be framed as (M)ILPs. Note that \ouracronym{} can be applied to (M)ILP problems without modification, because it makes no assumption about the optimization problem's structure. 
Very few methods are capable of applying the DFL paradigm to predict the constraint parameters of (M)ILP problems. As a baseline representative of this class, we consider \textsc{CombOptNet} \citep{paulus2021comboptnet}.

To evaluate \ouracronym{} for predicting parameters of constraints in an ILP problem, we considered two setups: the KP with unknown item weights, and the weighted set multi-cover (WSMC) with unknown coverage requirements. The mathematical models for KP and WSMC are provided in Appendix \ref{appendix:math_models}.
To introduce stochasticity, we use the  mapping described by \citet{elmachtoub2022smart} in their shortest path experiment.
We set the degree of model misspecification $deg=5$, the number of input features $p=5$, and the noise half-width $\bar{\epsilon} = 0.5$. The output of this mapping is then used to parameterize a Poisson distribution, both for the item weights in the KP, and the coverage requirements in the WSMC.
For the WSMC, the availability matrices were generated following a set of guidelines by \citet{grossman1997computational} that lead to realistic instances.
The set costs are generated uniformly at random from the range $\left[ 1, 100 \right]$.
We generate five datasets and for each dataset we consider three random splits, and use the same hyperparameter configurations as before.
As the relation between features $x$ and problem parameters $y$ is stochastic, for PFL, we switch to predicting a probabilistic model, allowing us to sample multiple scenarios; this model is trained by assuming that the parameters are normally distributed with non-contextual standard deviation, to mimic the fact that knowledge over the ground truth distribution is typically not available in practice. This probabilistic model is trained by minimizing the negative log-likelihood, without considering the task loss.

For the KP, the recourse actions allow for adding new items and discarding previously selected items. Given a penalty coefficient $\rho$, the value of a newly added item is $\frac{v}{\rho}$, while discarding a previously selected item incurs a cost of $\rho v$.
For the WSMC, the recourse action consists of adding extra units of the non-covered items at the price of paying an additional cost. The additional cost for each unit of unsatisfied coverage requirement of the $i$-th item is computed as the maximum set cost among the ones that cover it, multiplied by the coefficient $\rho$. 
We run experiments on the KP-50 and with $\rho \in \{ 5, 10, 20 \}$, and on the WSMC with 10 items and 50 sets. In \Cref{appendix:supp_res}, we provide additional results for the WSMC $5 \times 25$ and for a KP-50 with stochastic capacity instead of item weights, which lead to similar results.

\begin{table*}
    \centering
    \caption{Q1: Predicting Constraints Parameters - Integer linear programs. \textsc{PFL} and \ouracronym{} results on the KP-50 with uncertain weights. We omit the \textit{Feas. rel. PRegret} for \textit{Infeas. ratio}'s near 1.}
    \label{table:stochastic_weights_kp}
    \addtolength{\tabcolsep}{2pt}
    \small
    \begin{tabular}{lccccc}
    \toprule
    \textit{Method} & \textit{Rel. PRegret} & \textit{Feas. rel. PRegret} & \textit{Infeas. ratio} & \textit{MSE} & \textit{Epochs} \\
    \midrule
    
        \multicolumn{6}{c}{50-items, $\rho=5$} \\
        \midrule
        PFL                 & $0.168 \pm 0.036$           & $\mathbf{0.001 \pm 0.001}$ & $0.93 \pm 0.03$ & $\mathbf{7.88 \cdot 10^4 \pm 4.04 \cdot 10^4}$ & $\mathbf{34.0 \pm 20.3}$ \\
        \textsc{CombOptNet} & $0.189 \pm 0.068$           & $0.008 \pm 0.005$ & $\mathbf{0.91 \pm 0.02}$ & $5.26 \cdot 10^7 \pm 2.26 \cdot 10^7$          & $41.0 \pm 13.4$ \\
        SFGE(ours)          & $\mathbf{0.126 \pm 0.015}$  & -                 & $0.98 \pm 0.02$ & $3.62 \cdot 10^5 \pm 5.34 \cdot 10^4$           & $136.0 \pm 10.8$ \\
        \midrule
        
        \multicolumn{6}{c}{50-items, $\rho=10$} \\
        \midrule
        PFL                 & $0.319 \pm 0.081$           & $\mathbf{0.001 \pm 0.001}$ & $0.93 \pm 0.03$   & $\mathbf{7.88 \cdot 10^4 \pm 4.04 \cdot 10^4}$ & $\mathbf{34.0 \pm 20.3}$ \\
        \textsc{CombOptNet} & $0.400  \pm 0.146$          & $0.008 \pm 0.005$ & $\mathbf{0.91 \pm 0.02}$   & $5.26 \cdot 10^7 \pm 2.26 \cdot 10^7$          & $41.0 \pm 13.4$ \\
        SFGE(ours)          & $\mathbf{0.178 \pm 0.019}$  & -                 & $0.99 \pm 0.01$   & $3.71 \cdot 10^5 \pm 6.74 \cdot 10^4$          & $128.8 \pm 19.2$ \\

        \midrule
        
        \multicolumn{6}{c}{50-items, $\rho=20$} \\
        \midrule
        PFL                 & $0.615 \pm 0.174$          & $\mathbf{0.001 \pm 0.001}$ & $0.93 \pm 0.03$   & $\mathbf{7.88 \cdot 10^4 \pm 4.04 \cdot 10^4}$ & $\mathbf{34.0 \pm 20.3}$ \\
        \textsc{CombOptNet} & $0.822 \pm 0.302$          & $0.008 \pm 0.005$ & $\mathbf{0.91 \pm 0.02}$   & $5.26 \cdot 10^7 \pm 2.26 \cdot 10^7$          & $41.0 \pm 13.4$ \\
        SFGE(ours)          & $\mathbf{0.212 \pm 0.022}$ & -                 & $0.99 \pm 0.01$   & $3.73 \cdot 10^5 \pm 6.53 \cdot 10^4$          & $119.3 \pm 26.1$ \\

    \bottomrule
    \end{tabular}
\end{table*}

\begin{table*}[tb]
\centering
\vspace{0.5\baselineskip}
\caption{Q1: Predicting Constraints Parameters - Integer linear programs. PFL and \ouracronym{} results on the WSMC-$10 \times 50$. We omit the \textit{Feas. rel. PRegret} for \textit{Infeas. ratio}'s near 1.}
\label{table_main:wsmc}
\addtolength{\tabcolsep}{4pt}
\small
    \begin{tabular}{lccccc}
    
    \toprule

    \textit{Method}  &  \textit{Rel. PRegret}  & \textit{Feas. rel. PRegret}  &  \textit{Infeas. ratio}  &  \textit{MSE}         &  \textit{Epochs} \\
    
    \midrule

    \multicolumn{6}{c}{$10 \times 50$, $\rho=1$} \\

    \midrule
        
        \textsc{PFL}  & $2.35 \pm 0.85$           & $ - $                                    & $0.98 \pm 0.01$            & $ \mathbf{1.78 \cdot 10^5 \pm 3.03 \cdot 10^4} $                   & $\mathbf{35.3 \pm 44.2}$ \\
        \textsc{CombOptNet}                        & $ 3.04 \pm 1.20 $                       & $ - $                      & $ 1.0 \pm 0.0 $                                                    & $ 8.09 \cdot 10^6 \pm 5.24 \cdot 10^6 $                              & $ 49.9 \pm 29.5  $ \\
        \textsc{SFGE} (ours) & $\mathbf{1.93 \pm 0.50}$  & $ - $                                    & $\mathbf{0.94 \pm 0.05}$   & $ 3.60 \cdot 10^5 \pm 5.88 \cdot 10^4 $ 
                      & $70.3 \pm 13.1$ \\

    \midrule

    \multicolumn{6}{c}{$10 \times 50$, $\rho=5$} \\

    \midrule
        
        \textsc{PFL}  & $12.20 \pm 4.73$           & $\mathbf{0.034 \pm 0.019}$              & $0.96 \pm 0.01$            & $ \mathbf{2.01 \cdot 10^5 \pm 3.55 \cdot 10^4} $                    & $61.5 \pm 82.4$ \\
        \textsc{CombOptNet}                       & $ 88.80 \pm 34.3 $                       & $ - $                      & $ 1.0 \pm 0.0 $                                                     & $ 6.34 \cdot 10^6 \pm 2.64 \cdot 10^6 $                             & $ \mathbf{45.8 \pm 13.9} $ \\
        \textsc{SFGE} (ours) & $\mathbf{4.86 \pm 1.15}$  & $0.665 \pm 0.315$                       & $\mathbf{0.65 \pm 0.08}$   & $ 5.54 \cdot 10^5 \pm 1.25 \cdot 10^5 $                             & $81.4 \pm 16.9$ \\

    \midrule

    \multicolumn{6}{c}{$10 \times 50$, $\rho=10$} \\

    \midrule

        \textsc{PFL}  & $22.40 \pm 7.90$           & $\mathbf{0.027 \pm 0.041}$              & $0.98 \pm 0.01$            & $ \mathbf{2.18 \cdot 10^5 \pm 7.40 \cdot 10^4} $                    & $72.3 \pm 97.5$ \\
        \textsc{CombOptNet}                       & $ 374.17 \pm 79.0 $                     & $ - $                      & $ 1.0 \pm 0.0 $                                                         & $ 9.10 \cdot 10^6 \pm 3.95 \cdot 10^6 $                         & $\mathbf{34.0 \pm 15.7}$ \\
        \textsc{SFGE} (ours) & $\mathbf{7.08 \pm 1.29}$  & $1.30 \pm 0.53$                         & $\mathbf{0.54 \pm 0.14}$   & $ 7.39 \cdot 10^5 \pm 2.21 \cdot 10^5 $ 
                      & $67.7 \pm 14.4$ \\

    \bottomrule
    \end{tabular}
\end{table*} 

We can see in  \Cref{table:stochastic_weights_kp} and \Cref{table_main:wsmc} that
\textit{\ouracronym{} significantly outperforms the other methods in terms of expected relative post-hoc regret}. In the WSMC, \ouracronym{} consistently produces the lowest relative post-hoc regret. As $\rho$ increases, it becomes more conservative, reducing the infeasibility ratio, but resulting in a higher relative regret for solutions that do not require the recourse action. In the case of the KP, the \ouracronym{} solutions generally require recourse actions; but still consistently achieve the lowest post-hoc regret.
In terms of MSE, PFL is the most accurate, while \textsc{CombOptNet} is the least accurate.
Similarly to the previous experiments, \ouracronym{} 
converges slowly, requiring a relatively large number of epochs.


\subsection{Q2: Two-Stage Stochastic Optimization}

The ILP problems tackled in the previous section involve stochasticity in the ground-truth relation from features to problem parameters. 
Thus, the performance of the PFL method can be 
further improved by performing SAA over the predicted distribution \textit{at inference time} \citep{kleywegt2002sample}. This involves collecting a set of instance-specific samples, which are used as scenarios in the SAA algorithm to compute the optimal solution $z^\star$. The resulting solution is then employed to calculate the post-hoc regret. We refer to this pipeline as \textsc{PFL+SAA}.
In contrast, \ouracronym{} relies on stochasticity to smooth the regret, and does not aim to accurately learning the underlying distribution; moreover, our approach relies on \emph{a single sample} during inference. We compare these two methods to explore the scalability advantages of \ouracronym{} at inference time, and compare the quality of solutions obtained by both approaches. 

\begin{figure}
    \centering
    \includegraphics[width=0.85\textwidth]{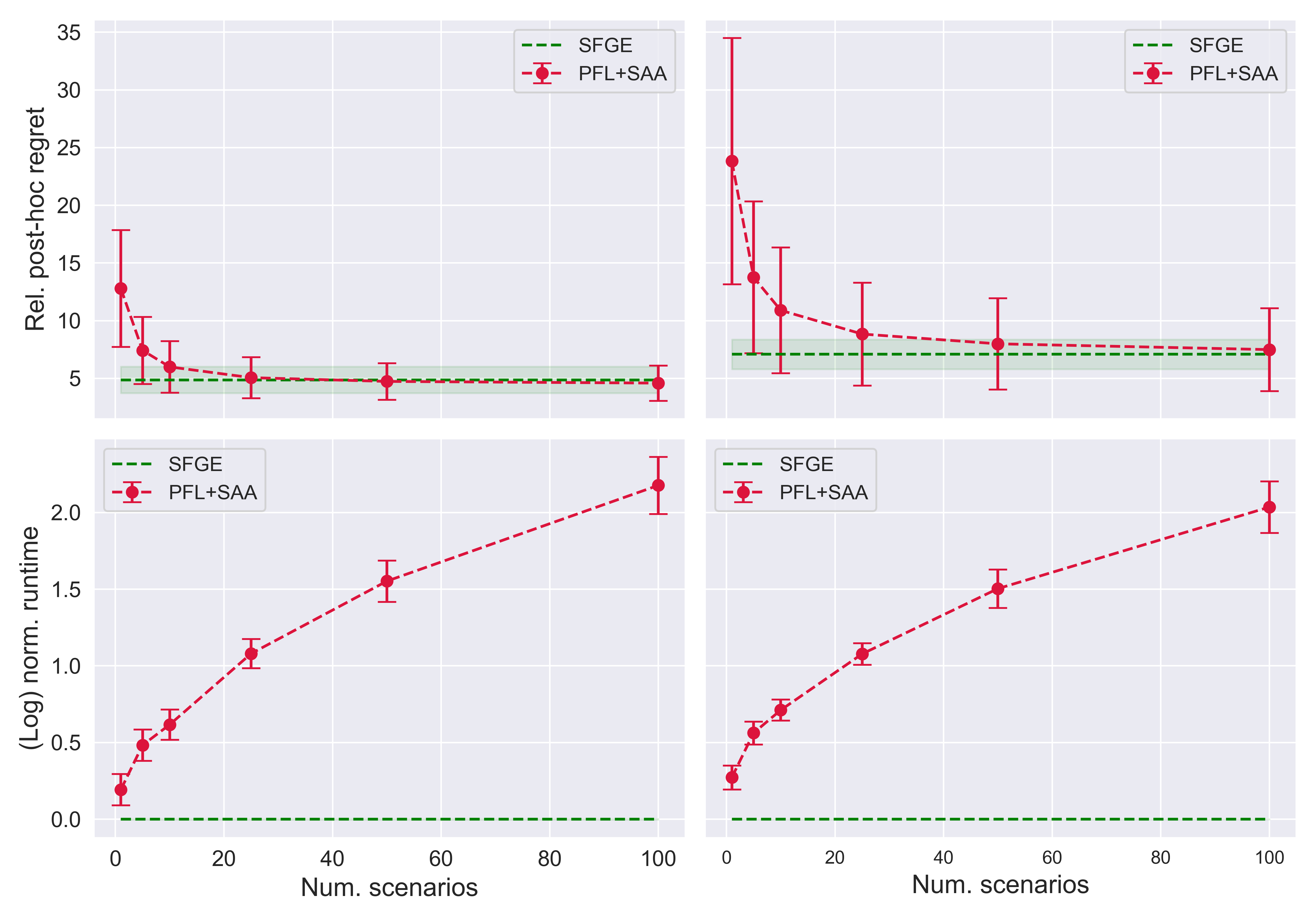}
    \caption{Q2: Two-Stage Stochastic Optimization. The relative post-hoc regret and normalized runtime at inference time of \ouracronym{} and PFL+SAA on the WSMC (size $10\times50$), for $\rho=5$ (left) and $\rho=10$ (right).}
    \label{fig:wsmc_10_50}
    \Description{The image is divided in 4 quadrants. A dashed curve for PFL+SAA and an horizontal line for SFGE. Standard deviation is displayed as vertical bars for PDF+SAA curve. Standard deviation is displayed as a trasparency area for SFGE.}
\end{figure}

\begin{figure}
    \centering
    \includegraphics[width=0.85\textwidth]{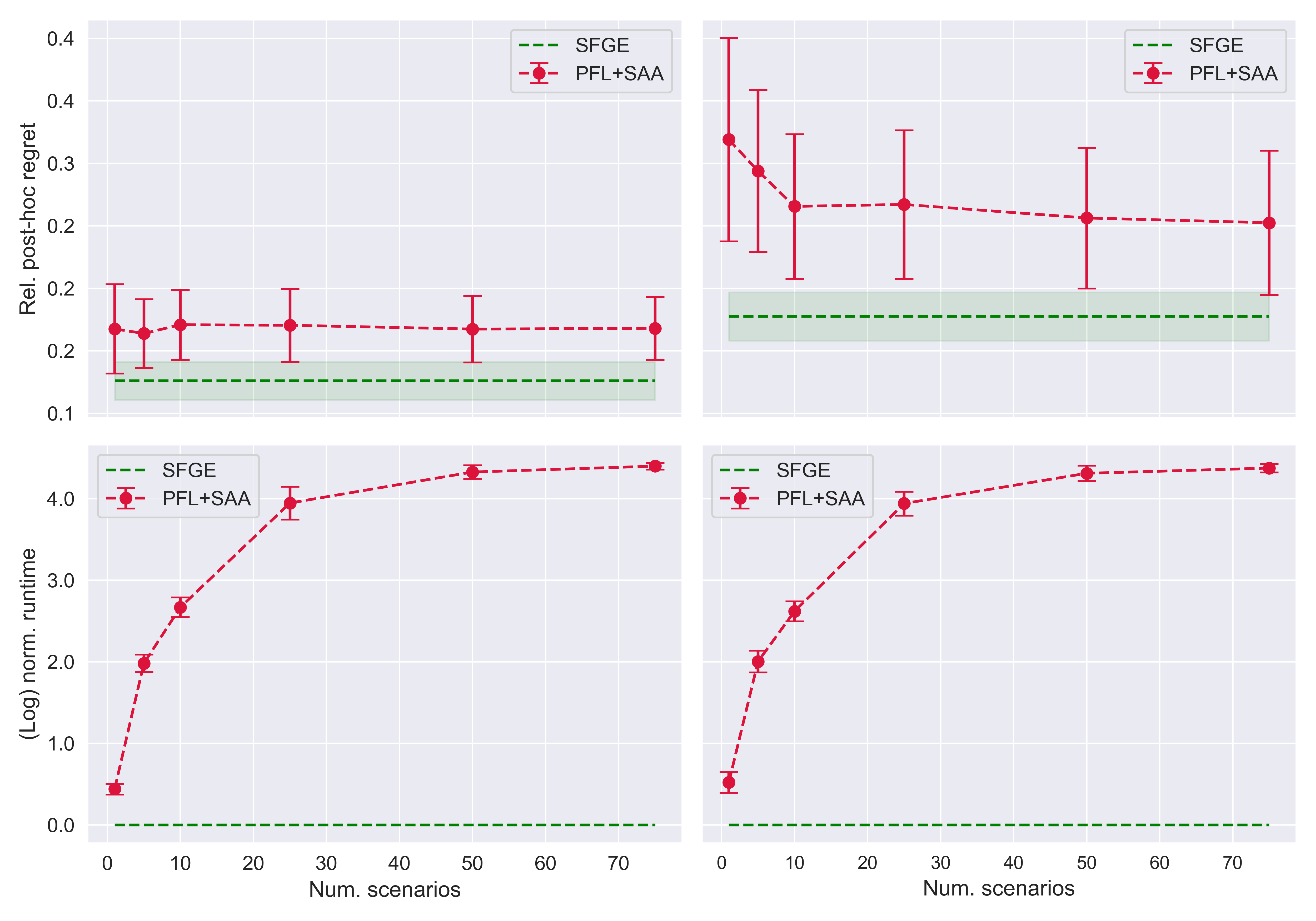}
    \caption{Q2: Two-Stage Stochastic Optimization. Comparison between \ouracronym{} and PFL+SAA on KP-5, for $\rho=5$ (left) and $\rho=10$ (right).}
    \label{fig:stochastic_weights_kp_50_items}
    \Description{The image is divided in 4 chars. A dashed curve for PFL+SAA and a horizontal line for SFGE. Standard deviation is displayed as vertical bars for PDF+SAA curve. Standard deviation is displayed as a trasparency area for SFGE.}
\end{figure}

In \Cref{,fig:wsmc_10_50,fig:stochastic_weights_kp_50_items}, we present the relative post-hoc regret (top row) and the normalized runtime during inference
on a logarithmic scale
(bottom row) against the number of sampled scenarios on the same ILP problems as before.
For the WSMC we could solve the optimization problem (with scenarios) to optimality; in the case of the KP, we had to impose a time limit of 30 seconds.
The corresponding values for \ouracronym{} are drawn as horizontal lines since \ouracronym{} does not sample scenarios. We can observe that with an increase in the number of scenarios, \textsc{PFL+SAA} generally improves on \textsc{PFL} in terms of relative post-hoc regret, but requires more computation time.
We also observe that for high $\rho$ values, \textit{\textsc{PFL+SAA} struggles to catch up to \ouracronym{}}.
Despite collecting 100 samples for KP and 75 samples for WSMC, \textsc{PFL+SAA} does not outperform \ouracronym{}. This is likely due to the model misspecification present in the experiment, as well as the fact that PFL relies on an assumed data distribution that differs from the ground-truth distribution (Gaussian vs. Poisson).
Additional results that support our conclusions are provided in \Cref{appendix:supp_res}.
These include, for all benchmarks involving a recourse action, an evaluation performed by sampling multiple realizations for every instance in the dataset.
The additional results confirm the trends observed above.

%


\subsection{Q3: Predicting Objective Parameters}

In the following experiment, we consider the task of predicting parameters that appear linearly in the objective function, a setting that most existing DFL methods have focused on. 
Since these methods are designed for this task, we do not expect \ouracronym{} to surpass them. Our aim is to assess how well \ouracronym{} performs compared to them.

We use the 0-1 KP with 50 items with unknown item values. We generate synthetic data by using the same mapping between input features and targets as described in the previous section, along with the same evaluation procedure. We again use the data generation process from \citet{elmachtoub2022smart} to introduce randomness into how features are mapped to item values.
We generate five datasets and for each dataset we consider three random splits. We use the same hyperparameter configurations as before.

In this setup, the SPO+ \cite{elmachtoub2022smart} and DPO (the Fenchel-Young alternative) \cite{berthet2020learning} losses will be used as a reference DFL methods for comparison: SPO+ has shown great performance across the board in existing comparative analyses~\cite{tang2022pyepo, mandi2023decision} while DPO relies on smoothing through perturbation, as our method does. 
When applying DPO we rely on a Gumbel distribution to perturb the solution with a scale factor $\epsilon=0.1$, and we approximate $dz^{\star}/dy$ with a single Monte Carlo sample.
We also employ a baseline PFL model, trained to minimize the MSE between the predictions and targets.

Convergence speed is important for DFL methods, since for each gradient update step we need to solve an optimization problem, which can be computationally expensive thus hindering its application to real-world scenarios. To improve convergence speed, inspired by \citet{mulamba2020contrastive}, we propose an alternative task loss for problems where predictions occur linearly in the objective function. 
The task loss is:
\begin{equation}
    \mathcal{L}(\hat{y}, y) = Regret(\hat{y}, y) + f(\hat{y}, z) - f(\hat{y}, \hat{z})
\end{equation}
where $f$ is the objective function. We add the regret with respect to the predicted solution to the original task loss (the $\mathcal{L}_{MAP}$ as described in \citet{mulamba2020contrastive}).
We refer to this method as \ouracronym{}-MAP.

\begin{table}
\centering
\caption{Q3: Predicting Objective Parameters. PFL, \textsc{\ouracronym{}} and \textsc{SPO} results on the KP-50 with uncertain objective coefficients.}
\label{table_main:knapsack}
\small  
    \begin{tabular}{lccc}
    
    \toprule
    
    \textit{Method} & \textit{Rel. regret} & \textit{MSE} ($\times 10^4$) & \textit{Epochs} \\


    
    \midrule
        
        \textsc{PFL}      & $ 0.022 \pm 0.005 $          & $ \mathbf{2.38  \pm 1.28  } $           & $ \mathbf{40.5 \pm 28.5} $ \\
        \textsc{SPO}      & $ 0.003 \pm 0.0004 $ & $ 4.12 \pm 1.52 $                       & $ \mathbf{40.3 \pm 9.77} $ \\
        \textsc{DPO}      & $ \mathbf{0.002 \pm 0.001} $ & $ 6.24 \pm 3.30 $                       & $ 45.0 \pm 13.4 $ \\
        \ouracronym{}     & $ 0.009 \pm 0.002 $          & $ 17.5  \pm 3.34  $                     & $ 149.8 \pm 15.1 $ \\
        \ouracronym{}-MAP & $ 0.008 \pm 0.002$           & $ 10.7 \pm 2.03 $                       & $ 83.1 \pm 8.03 $ \\


    
        
    \bottomrule
    \end{tabular}
\end{table} 

The aggregated results are reported in \Cref{table_main:knapsack}.
Looking at the relative regret, we see that SPO performs best, but that \ouracronym{} is still able to outperform PFL significantly.
In terms of convergence speed, SPO and PFL require a comparable number of epochs whereas \ouracronym{} is significantly slower to converge. We also observe that \ouracronym{}-MAP improves convergence speed by a noticeable margin and should therefore, when applicable, be preferred over its standard counterpart.

Moreover, in \Cref{table:knapsack} of \Cref{appendix:supp_res} we present additional results for the KP-50 dataset and for quadratic KP instances, with similar outcomes.
In conclusion, it appears that \ouracronym{} may not be the best choice when the uncertain parameters represent coefficients of a linear cost function; existing DFL methods can effectively address this particular problem in less training epochs.
However, the fact that \ouracronym{} outperforms PFL also in this case allows us to position SFGE as a generic DFL method, capable of tackling a wide range of predict-then-optimize problems.


\subsection{Q4: Widening Applicability}

Existing approaches are limited by the specific assumptions they make on the optimization problem's structure. Conversely, \ouracronym{} makes no assumptions about problem structure, and can thus straightforwardly be applied to any setting, and can be combined with any (black-box) solver.

To evidence this advantage, we performed an experimental evaluation on a problem for which, to the best of our knowledge, no other DFL approach can be successfully applied, showing how \ouracronym{} can be the only candidate solution to surpass PFL in certain use cases.
Concretely, we considered a production planning problem characterized by a quadratic objective function. Given a set of $n$ products $p_1, p_2, \dots, p_n$, each with an associated overproduction and underproduction cost $o_i$ and $u_i$, and a maximum capacity $c$, the goal is to find the optimal integer production values $v \in \mathbb{Z}_+^n$ to minimize the overall cost, computed as:
\begin{equation*}
    \text{f}(z, y) = \sum_{i=1}^n o_i \max(0, z_i - y_i)^2 + u_i \max(0, y_i - z_i)^2,
\end{equation*}
with $y \in \mathbb{Z}_+^n$ being the (predicted) products demands. This combinatorial problem does not offer informative gradients, due to the integrality constraints. In addition, the nonlinearity of the objective function does not allow the use of methods utilizing an LP relaxation (e.g., \citet{hu2023predict+}).

To generate the dataset, we sample input features $x \in \mathbb{R}^m$ and a matrix of weights $W \in \mathbb{R}^{n \times m}$ from bounded uniform distributions. Then, given $k$ potential customers, we compute ground-truth demands as:
\begin{equation*}
    y \sim Bin(k, \sigma(Wx)),
\end{equation*}
with $\sigma$ being the sigmoid function. This process defines a stochastic setting that correlates demands and observable features. As a simple yet demonstrative case, we set $n = 10$, $m = 4$, $k=100$, $c=400$. We create an asymmetry between overproduction and underproduction costs by sampling $u$ from $[0.05, 0.3] \cup [0.7, 0.95]$ and setting $o = 1 - u$.

\begin{table}[tb]
\centering
\caption{Q4: Widening Applicability. PFL and \textsc{\ouracronym{}} results on the production planning problem.}
\label{table_main:production_planning}
\small  
    \begin{tabular}{lccc}
    
    \toprule
    
    \textit{Method} & \textit{Rel. regret} & \textit{MSE} & \textit{Epochs} \\
    
    \midrule
        
        \textsc{PFL}  & $ 211.26 \pm 33.01 $ & $ \mathbf{21.44  \pm 0.57  } $ & $ \mathbf{142.16 \pm 24.12} $ \\
        \ouracronym{} & $ \mathbf{116.04 \pm 14.35} $ & $ 39.65  \pm 0.15  $ & $ 246.08 \pm 12.53 $ \\
    \bottomrule
    \end{tabular}
\end{table} 
Results are reported in \Cref{table_main:production_planning}. \ouracronym{} significantly outperforms PFL in terms of regret, as PFL converges towards the mean value of $y$ given $x$, which does not lead to optimal decision quality. In contrast, \ouracronym{} learns to appropriately underestimate or overestimate demands, according to the corresponding underproduction and overproduction costs.

\section{Conclusions and Future Directions}
\label{sec:conclusions}
This work widens the applicability of DFL  by proposing a method that \emph{does not assume structural properties of the task at hand}. 
Concretely, we employ stochastic smoothing and \ouracronym{} to estimate the gradients of a smoothed loss. 
This allows the method to be applied to linear and nonlinear problems, with or without integrality constraints, and with uncertainty in the objective, in the constraints, or in both. As a specific use case, we show how our method can be used to address two-stage stochastic problems by using only a single predicted parameter vector and a deterministic optimization model, without the need for SAA.  
Our experimental evaluation reveals that, for problems with uncertainty in the constraints, \ouracronym{} matches or outperforms existing methods in terms of post-hoc regret.
When predicting parameters that appear linearly in the objective function, \ouracronym{} does not outperform the state of the art, but still provides a major improvement over PFL approaches.
The main drawback of our method is that it tends to be slower in convergence speed compared to existing methods. This can partly be attributed to the high variance in the Monte Carlo gradient estimates. We alleviate this drawback by standardizing the regret on each mini-batch, as a way to reduce variance.

An interesting direction for future work is to employ SFGE on a real-world case study. The experiments in this paper were conducted on well-defined synthetic benchmarks from the literature to allow for a controlled evaluation. In practice, real datasets often offer weaker signal, more noise, and may lead to higher variance in the SFGE estimator. To account for this, alternative variance reduction techniques from the literature can be used, for example, employing an actor-critic style algorithm. Another interesting direction is to investigate the merit of other Monte Carlo gradient estimators, such as the measure-valued gradient estimator, as an alternative to \ouracronym{}.



\section*{Acknowledgments}
This research received funding from the European Research Council (ERC) (Grant No. 101002802, CHAT-Opt), from the European Union's HORIZON-CL4--2021-HUMAN-01 research and innovation programme (Project Tuples, Grant Agreement No. 101070149), and from the Research Foundation Flanders (FWO) project G0G3220N. Senne Berden is a fellow of the Research Foundation -- Flanders (FWO-Vlaanderen, 11PQ024N).

\clearpage

\bibliographystyle{icml2024}
\bibliography{sample-base}

\newpage
\appendix
\onecolumn
\setcounter{secnumdepth}{1} 

\section{LP Relaxations May Fail to Guide DFL Methods towards an Optimum}
\label{app:lp_vs_milp}
We illustrate that replacing a MILP problem with its LP relaxation can cause a mismatch between the optimal predicted parameters for the true task loss and the corresponding approximation, as mentioned in \cref{sec:related_work}.
In particular, consider the following simple mixed-integer linear program:
\begin{align}
\min \ & y_0 z_0 + y_1 z_1  & (y)\\
\text{s.t.} \
& 2(u-\varepsilon)z_0 + z_1 \geq u - \varepsilon & (c_0) \\
& - 2\varepsilon z_0 + z_1 \geq - \varepsilon & (c_1) \\
& z_0 \geq 0 & (c_2) \\
& z_0 \leq 1 & (c_3) \\
& z_1 \leq u & (c_4) \\
& z_0 \in \mathbb{Z}
\end{align}
where the constraints are labeled as $c_0$, $c_1$, $c_2$, $c_3$ and $c_4$ and $\varepsilon > 0$ is a small real number.
Assume that the ground truth cost coefficients are given as $y = (0,1)$.

\begin{figure}[tbh]
\centering
\begin{subfigure}[t]{0.3\textwidth}
  \centering
  \includegraphics[width=\linewidth]{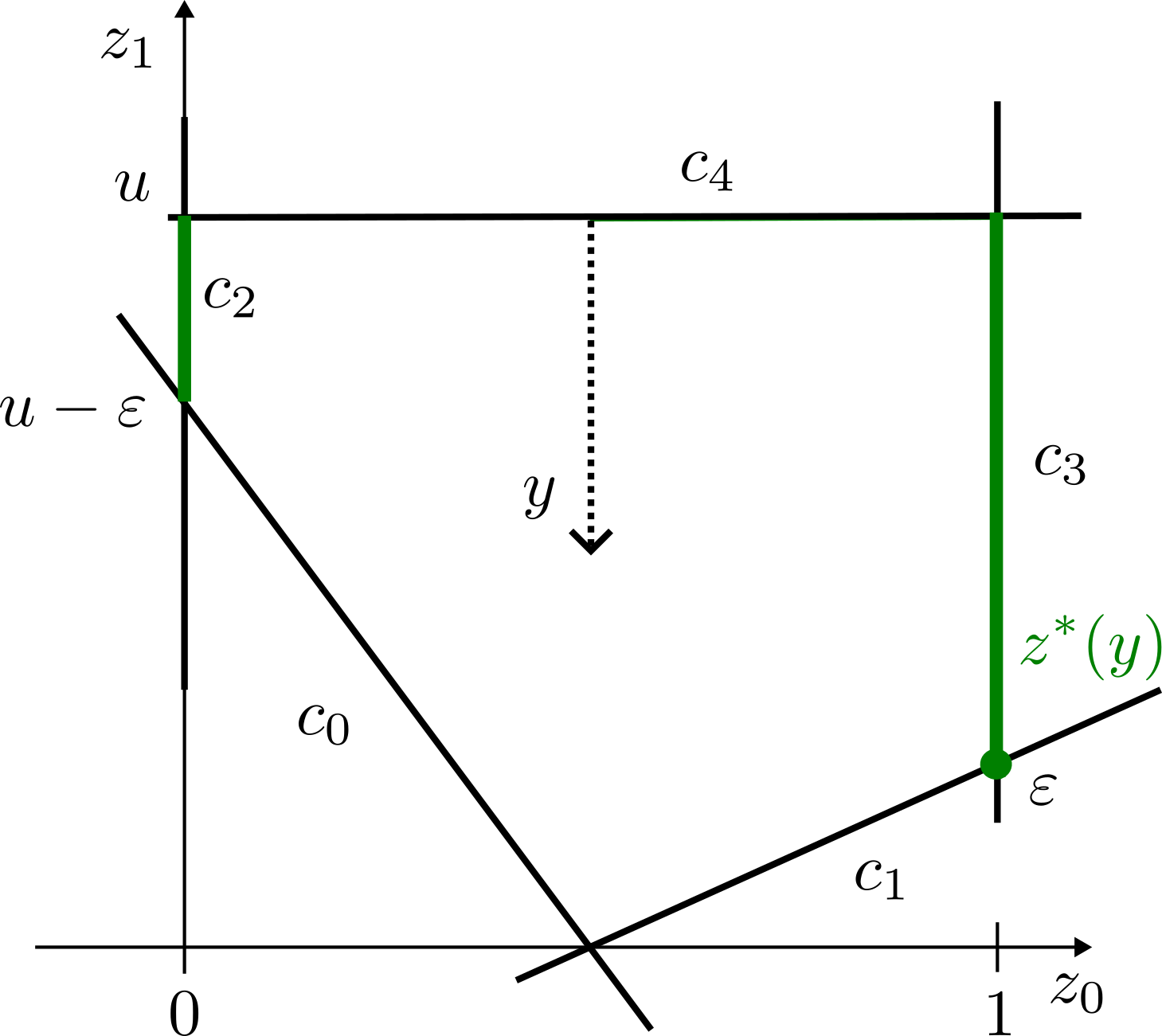}
  \caption{}\label{fig:ce_milp}
\end{subfigure}
\begin{subfigure}[t]{0.3\textwidth}
  \centering
  \includegraphics[width=\linewidth]{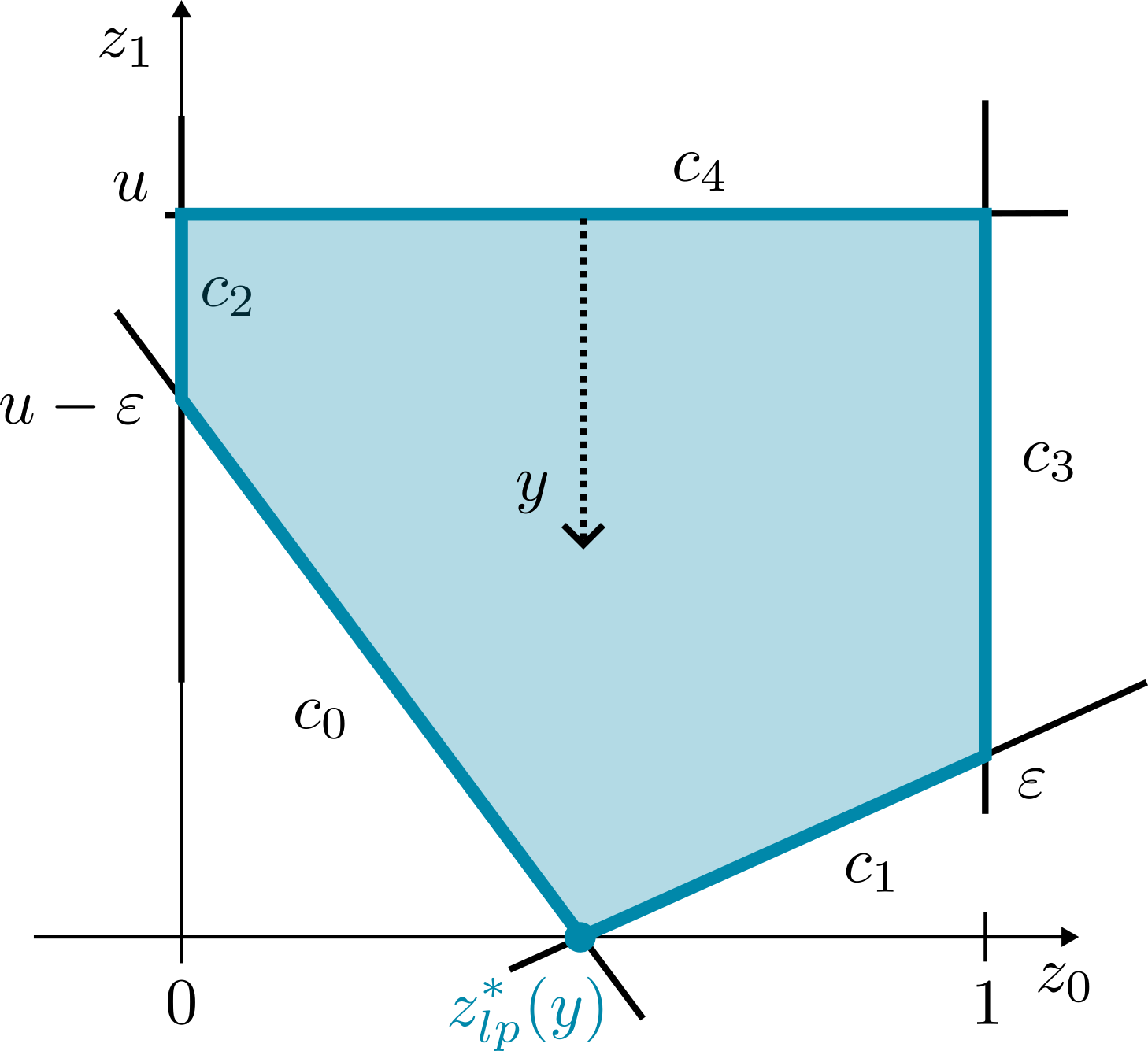}
  \caption{}\label{fig:ce_lp}
\end{subfigure}
\caption{Feasible space and optimum for the example MILP (\subref{fig:ce_milp}) and its LP relaxation (\subref{fig:ce_lp})}\label{fig:ce_problems}
\end{figure}

Note that this problem matches a traditional predict-then-optimize setup, with unknown parameters $y$ appearing in a linear cost function.
\Cref{fig:ce_milp} visually depicts the feasible space (in green) and the optimal solution with respect to the ground truth cost coefficients for the MILP. \Cref{fig:ce_lp} provides the same information (in blue) for the corresponding LP relaxation, obtained by dropping the integrality constraint.

When the ML model manages to estimate the ground-truth parameter vector $y$ correctly, the regret is 0 for both the MILP and the LP formulations, as expected.
However, an incorrect prediction may still lead to the correct optimal solution.

\begin{figure}[tbh]
\centering
\begin{subfigure}[t]{0.3\textwidth}
  \centering
  \includegraphics[width=\linewidth]{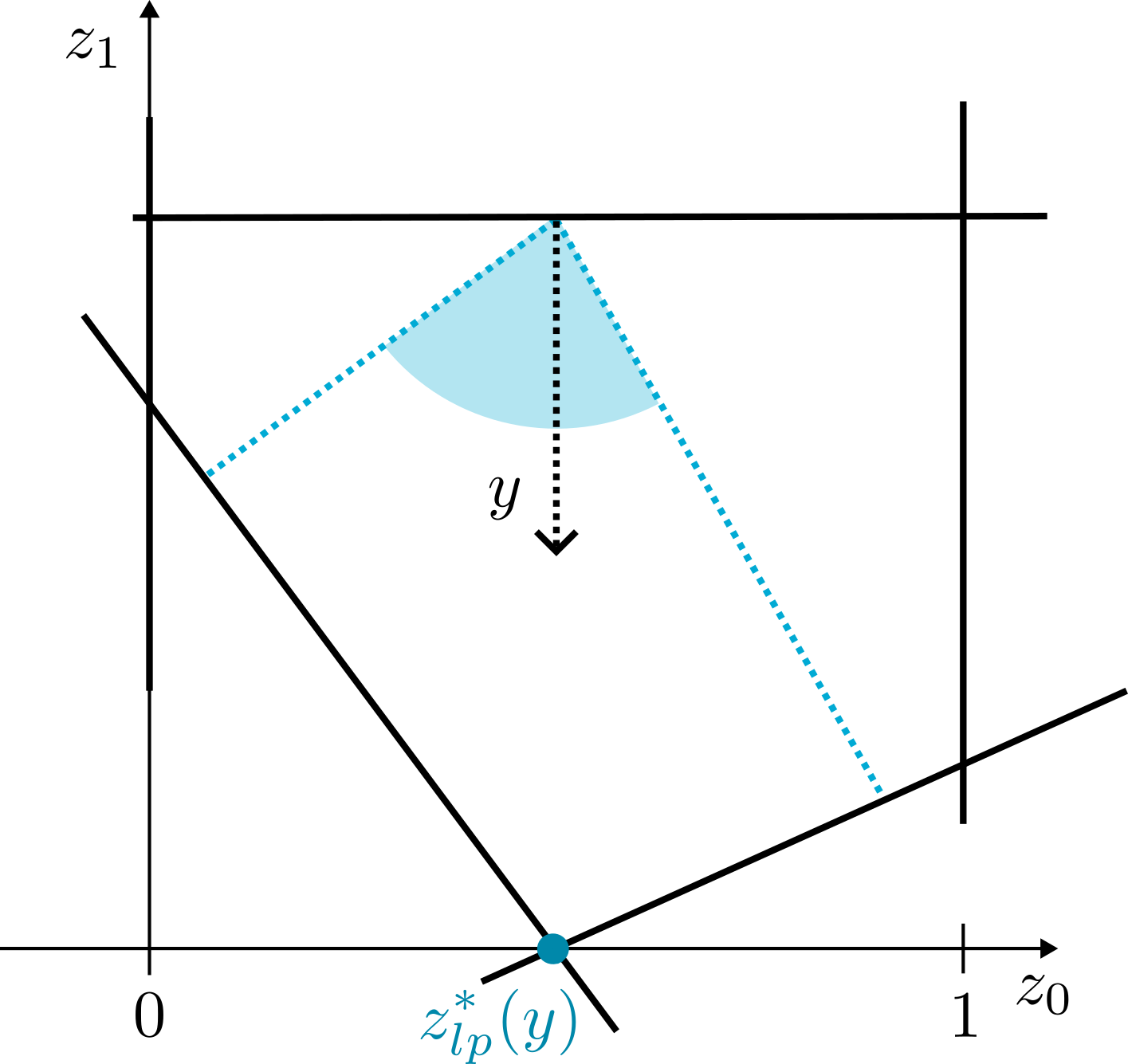}
  \caption{}\label{fig:ce_lp_grad}
\end{subfigure}
\begin{subfigure}[t]{0.3\textwidth}
  \centering
  \includegraphics[width=\linewidth]{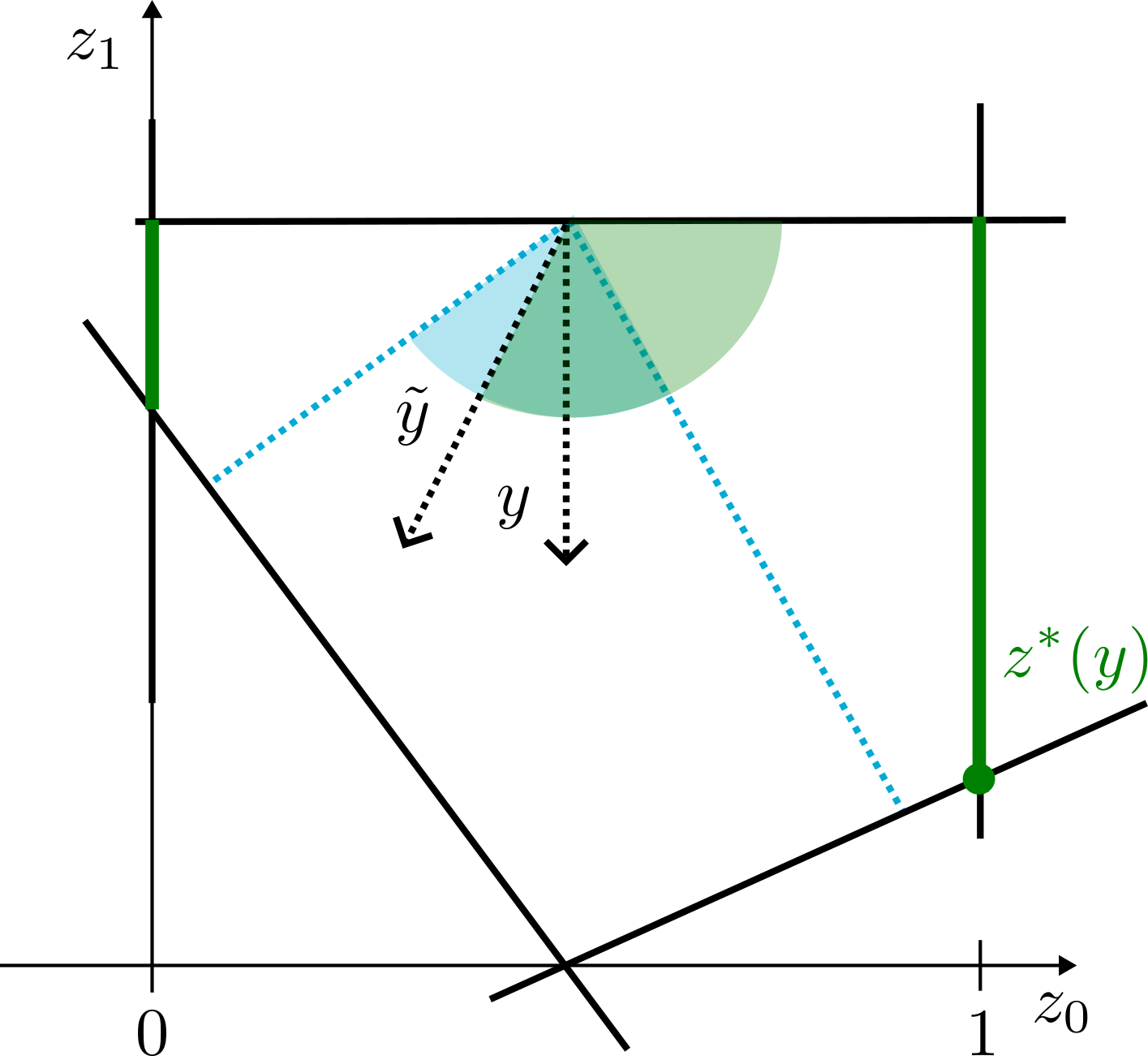}
  \caption{}\label{fig:ce_milp_grad_good}
\end{subfigure}
\begin{subfigure}[t]{0.3\textwidth}
  \centering
  \includegraphics[width=\linewidth]{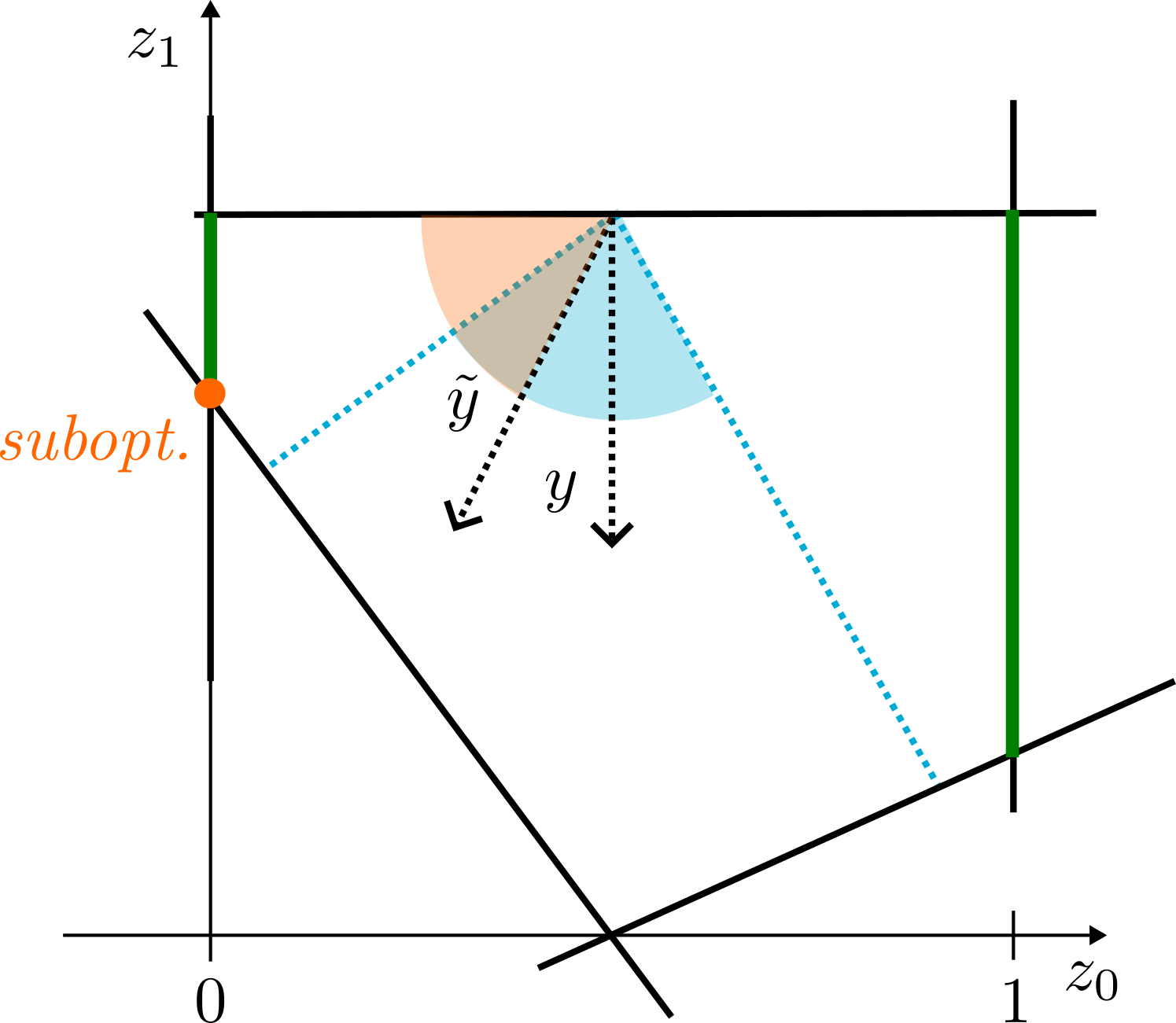}
  \caption{}\label{fig:ce_milp_grad_bad}
\end{subfigure}
\caption{(\subref{fig:ce_lp_grad}) Range of equivalent optimal predictions for the LP relaxations;
(\subref{fig:ce_milp_grad_good}) range of predictions leading to the correct MILP optimum;
(\subref{fig:ce_milp_grad_bad}) range of predictions leading to a suboptimal MILP solution although they are optimal for the LP relaxation.
}\label{fig:ce_gradients}
\end{figure}

\Cref{fig:ce_lp_grad} visually depicts the range of predictions that result in the correct LP optimum. If the ML model outputs any of these vectors, the regret (on the LP relaxation) will still be 0. \Cref{fig:ce_milp_grad_good} provides a similar analysis for the MILP formulation, i.e., the one representing the true task loss rather than an approximation. Any predicted vector with a direction on the right-hand side of $\tilde{y} = (u-2\varepsilon, 1)$ -- the green-shaded angle -- will lead to the correct optimal solution.

There is some overlap between the set of optimal MILP predictions and the set of optimal LP predictions.
The larger this overlap, the more effective the use of the LP relaxations in guiding DFL towards high-quality solutions.

However, as depicted in \Cref{fig:ce_milp_grad_bad}, the set of optimal LP predictions also includes vectors that result in a MILP solution with significantly higher cost. In the figure, the orange angle identifies the parameter vectors that result in an incorrect MILP optimum, which overlaps with the blue angle of optimal predictions for the LP relaxation. The larger this overlap, the less effective LP relaxations become.

\section{Mathematical Models of the Optimization Problems}
\label{appendix:math_models}

\paragraph{Mathematical Model of the Knapsack Problem} 
Given a set of items $\mathcal{I}$, let $w_i$ be the weight of item $i \in \mathcal{I}$. Also, let $v_i$ be the value of item  $i \in \mathcal{I}$. The Knapsack Problem (KP) aims to maximize the total value while ensuring that the total weight of selected items does not exceed a given capacity $W$. The following mathematical model solves the KP:
\begin{align}
    \max_{z} & \sum_{i \in \mathcal{I}} v_iz_i \label{eq:KPobj} \\
    \text{s.t.} &  \sum_{i \in \mathcal{I}} w_iz_i \leq W  \label{eq:KPcons}  \\
    &  z_i \in \{0,1\} \text{ for all }  i \in \mathcal{I} \label{eq:KPint}.
\end{align}
where the binary  variable $z_i$ indicates if item $i$ is selected. The objective \eqref{eq:KPobj} maximizes the value of selected items. Constraint \eqref{eq:KPcons} ensures that the capacity is respected. In the quadratic KP, the objective is replaced by $\sum_{i \in \mathcal{I}} \sum_{j \in \mathcal{I}}v_{ij}z_iz_j$. In the fractional version of the problem, the domain constraint $z_i \in \{0,1\}$ is replaced by $z_i \in [0,1]$.

\paragraph{Mathematical Model of the Stochastic Knapsack Problem} 

The KP with unknown items' weights is a 2-stage stochastic optimization problem. Given the predicted weights $\hat{w}$, we compute the optimal solution $\hat{z}$ by solving the optimization problem described in \cref{eq:KPobj,eq:KPcons,eq:KPint}. During the second stage, we need to find the optimal recourse actions that maximize the value of the selected items by solving the following optimization problem:
\begin{align} 
    \max_{u^+, u^-} \quad &  \sum_{i \in \mathcal{I}} \frac{1}{\rho} v_i u_i^+ - \rho v_i u_i^- \\ 
    \text{s.t.} \quad   & \sum_{i \in \mathcal{I}} w_i(\hat{z}_i + u_i^+ - u_i^-) \leq C \\
                        & \hat{z} \geq u^- \\
                        & \hat{z} + u^+ \leq 1 \\
                        & u^+, u^- \in \{0,1\}
\end{align}
where $u^+$ and $u^-$ are respectively the selected/removed items during the second stage, $w$ is the realization of the items' weights and $\rho > 1$ is the penalty coefficient.

The SAA involves solving the first and second stages in a single model. The first stage decisions are the same for all the scenarios, while we need a set of recourse actions for each scenario. The resulting model is:
\begin{align} 
\max_{z, u^+_{\omega}, u^-_{\omega}} \quad & \sum_{i \in \mathcal{I}} v_i z_i + \frac{1}{|\Omega|}\sum_{\omega \in \Omega} \frac{1}{\rho} v_i u^+_{i, \omega} - \frac{1}{|\Omega|}\sum_{\omega \in \Omega}\rho v_i u_{i, \omega}^- \\ 
\text{s.t.} \quad                 & \sum_{i \in \mathcal{I},\omega \in \Omega}w_{i,\omega}(z_i + u^+_{i,\omega} - u_{i,\omega}^-) \leq C \qquad & \forall \omega \in \Omega \\
                  & z \geq u_{\omega}^- & \forall \omega \in \Omega \\
                  & z + u^+_{\omega} \leq 1  \qquad & \forall \omega \in \Omega \\
                  & z, u^+_{\omega}, u_{\omega}^- \in \{0,1\}, \omega \in \Omega
 \end{align}
where $\omega \in \Omega$ are the sampled scenarios.
When the capacity is uncertain, the model is similar except for the fact that the weights $w$ are known and the capacity is sampled for each scenario.

\paragraph{Mathematical Model of Weighted Set Multi-Cover Problem} 
Let $\mathcal{I}$ be the set of items and $\mathcal{J}$ be the set of covers. The parameter $a_{ij}$ is 1 if $j$ can cover $i$ and 0 otherwise. Item $i \in \mathcal{I}$ must be covered at least $d_i$ times. The cost of selecting cover $j \in \mathcal{J}$ is $c_j$. The weighted set
multi-cover problem (WSMC) aims to satisfy coverage constraints while minimizing the total cost. The following mathematical model solves the WSMC:

\begin{align}
    \min & \sum_{j \in \mathcal{J}} c_jz_j \label{eq:WSMCobj} \\
    \text{s.t.} &  \sum_{j \in \mathcal{J}} a_{ij}z_j \geq d_i  & \quad \forall  i \in \mathcal{I} \label{eq:WMSCcons}  \\
    &  z_j \geq 0 \text { and integer}                          & \quad \forall  j \in \mathcal{J} \nonumber
\end{align}
where the non-negative integer variable $z_j$ indicates how many times cover $j$ is selected. The objective \eqref{eq:WSMCobj} minimizes the total cost. Constraint \eqref{eq:WMSCcons} ensures the coverage requirement for each item. 

\paragraph{Mathematical Model of the stochastic WSMC}
In our formulation of the stochastic WSMC, the items' coverage requirement is unknown at solution time, resulting in a two-stage stochastic optimization model that can be formulated as follows:
    \begin{align}
        \min & \sum_{j \in \mathcal{J}} c_j z_j + \sum_{i \in \mathcal{I}} \rho s_{i}  \\
             & \sum_{j \in \mathcal{J}} a_{i, j} z_j \geq d_{i} (1 - w_{i})               & \quad \forall i \in \mathcal{I} \\ 
        & w_{i} = 1 \implies s_{i}  \geq d_{i} - \sum_{j \in \mathcal{J}} a_{i, j} x_j    & \quad \forall i \in \mathcal{I} \nonumber  \\
        & z_j \geq 0  \\ 
        & w_{i} \in \left[ 0, 1 \right]  \\ 
        & s_{i} \geq 0  \\ 
        & z, w \in \mathbb{Z} 
    \end{align}
where $w$ are indicator variables and $s$ is a set of slack variables corresponding to the non-satisfied coverage requirements.

Similarly to the KP, we can obtain an SAA formulation by sampling the coverage requirements $d$ and introducing a set of slack variables for each scenario $\omega \in \Omega$:

    \begin{align}
        \min & \sum_{j \in \mathcal{J}} c_j z_j + \frac{1}{|\Omega|} \sum_{\omega \in \Omega} \sum_{i \in \mathcal{I}} \rho_{i, \omega} s_{i, \omega} \\
             & \sum_{j \in \mathcal{J}} a_{i, j} z_j \geq d_{i, \omega} (1 - w_{i, \omega})                      & \quad \forall i \in \mathcal{I}, \omega \in \Omega \nonumber \\ 
        & w_{i, \omega} = 1 \implies s_{i, \omega}  \geq d_{i, \omega} - \sum_{j \in \mathcal{J}} a_{i, j} x_j   & \quad \forall i \in \mathcal{I}, \omega \in \Omega \nonumber \\
        & z_j \geq 0 \\ 
        & w_{i, \omega} \in \left[ 0, 1 \right] \\ 
        & s_{i, \omega} \geq 0 \\ 
        & z, w \in \mathbb{Z} 
        \end{align}
%

\section{Methodological Details}
\label{appendix:distr_and_grad}

\paragraph{Non-contextual standard deviation}
\begin{figure}[htb]
\centering
\includegraphics[width=0.9\textwidth]{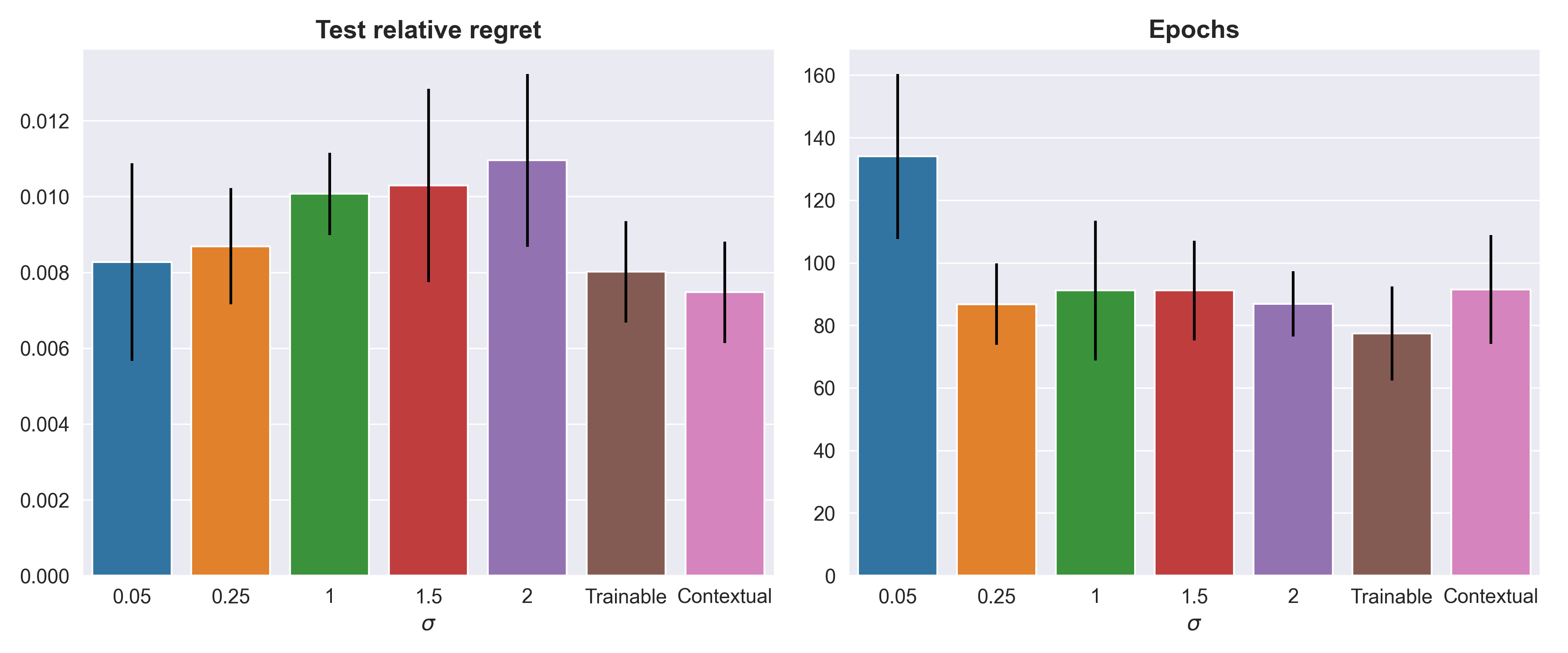}
\caption{Relative regret (left) and number of training epochs before early stopping (right) required by \ouracronym{} for different fixed values of $\sigma$, for a trainable $\sigma$ and for a contextual $\sigma$, on the KP-50. We employed the same data generation and evaluation procedures described in \Cref{sec:exp_res}.}
\label{fig:static_vs_fixed_std_dev}
\Description{The image is divided in two plots. Both contain barplots. The barplots height represents the relative regret for the left handside plot. The barplots height represents the number of epochs for the right handside plot.}
\end{figure}

While we employ stochastic parameter estimates, it is important to note that they are a part of our approach to smoothing and need not precisely reflect the actual distribution of $y$. Thus, we opt for a Gaussian distribution (defined through $\mu$ and $\sigma$), not because it most accurately represents the variance in $y$, but because it results in effective localized smoothing to obtain informative gradients. Throughout this research, we experimented with various ways of controlling the standard deviation $\sigma$, including the use of a constant hyperparameter $\sigma$, a trainable non-contextual $\sigma$ (i.e., which is not dependent on the input features), and a contextual $\sigma$ (i.e., which is input-dependent and is predicted alongside $\mu$, based on features $x$). We observed that using a trainable non-contextual standard deviation generally produced the best tradeoff in terms of relative regret and number of epochs required for convergence (as shown in \Cref{fig:static_vs_fixed_std_dev}). Introducing dependence on input features did not lead to a significant improvement, and thus, we use this setup in our experiments. This result intuitively makes sense, since $\sigma$ only controls the degree of smoothing, and is not used to model any true variance that may be present in the data (after all, at test time we do not sample from the predicted distribution, and instead use point prediction $\mathbb{E}_{\hat{y} \sim p_\theta}[\hat{y}] = \mu$ directly).

\paragraph{Number of samples}
\begin{figure}[htb]
\centering
\includegraphics[width=0.6\textwidth]{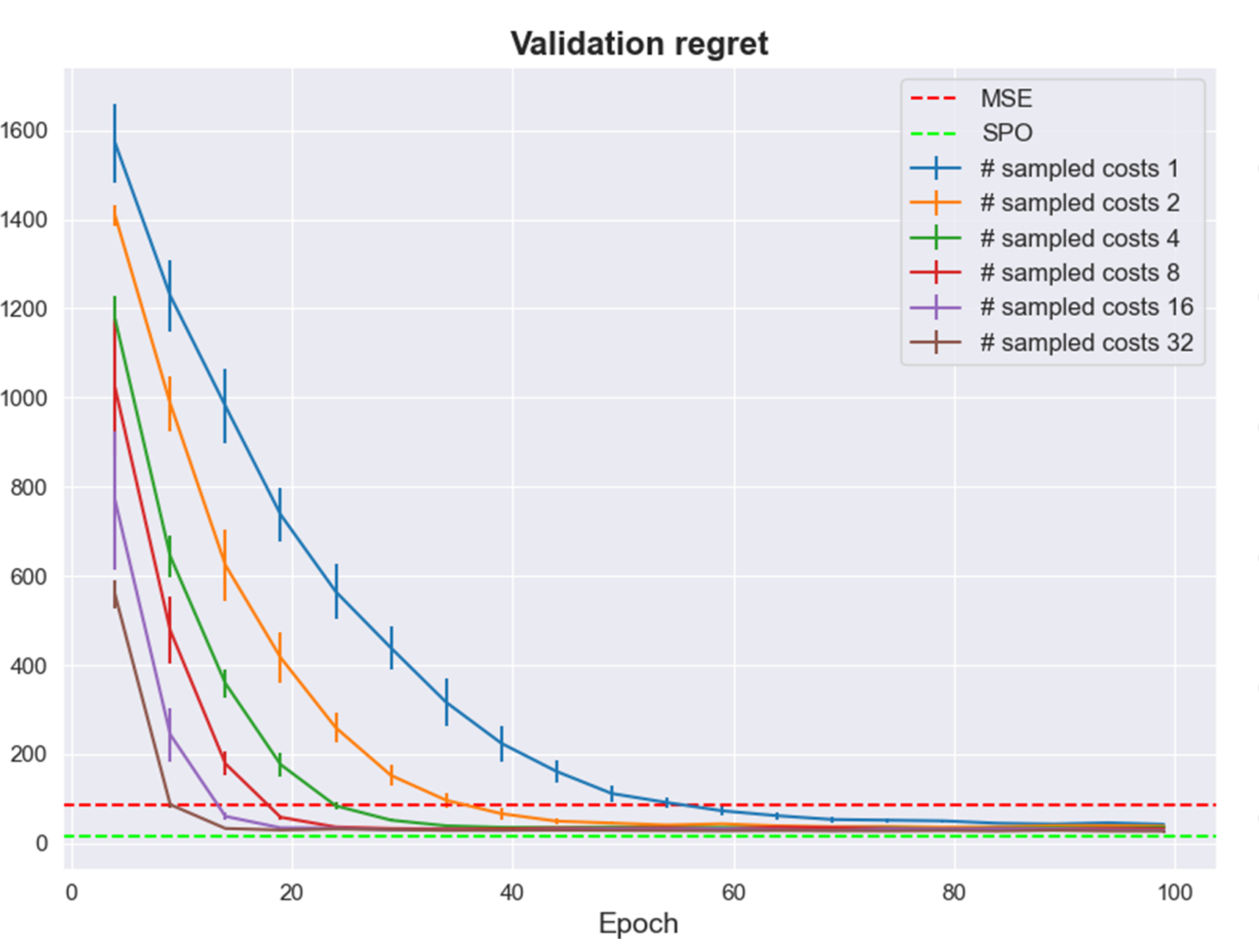}
\caption{The validation regret on the KP-50 with respect to the number of epochs, for different values of $S$ (denoting the number of predictions $\hat{y}$ that are sampled per instance in each gradient estimation).}
\label{fig:add_results}
\Description{The image shows one curve for each value of S. The value of S are 1, 2, 4, 8, 16 and 32. Two horizontal dashed lines are displayed referring to PDF and SPO methods.}
\end{figure}

The gradient estimator is given by:
\begin{align}
\begin{split}
    \nabla_{\theta} L(\theta, y) &\approx \frac{1}{S} \sum_{i=1}^{S} \mathcal{L}(\hat{y}^{(i)}, y) \nabla_{\theta} \log p_{\theta}(\hat{y}^{(i)}) \\
\end{split}
\end{align}
The motivation for using a higher number of samples is to obtain a better estimate of the true gradient, which can improve convergence speed. However, this also requires solving more optimization problems per gradient descent step, which may slow convergence speed in actual wall-clock time. We investigated this trade-off: as shown in \Cref{fig:add_results}, using more samples results in fewer training epochs. However, we found that in terms of number of optimization problems solved until convergence (and consequently, wall clock time), $S=1$ works best in our experiments. In general, the right value for $S$ is likely to depend on the problem. For problems where the gradient estimate has high variance and the optimization problem can be solved relatively quickly, larger values for $S$ may provide a better trade-off between the accuracy of the gradient estimate and the computational time required to compute it. 


\paragraph{Variance reduction}
\begin{figure}[htb]
\centering
\includegraphics[width=0.6\textwidth]{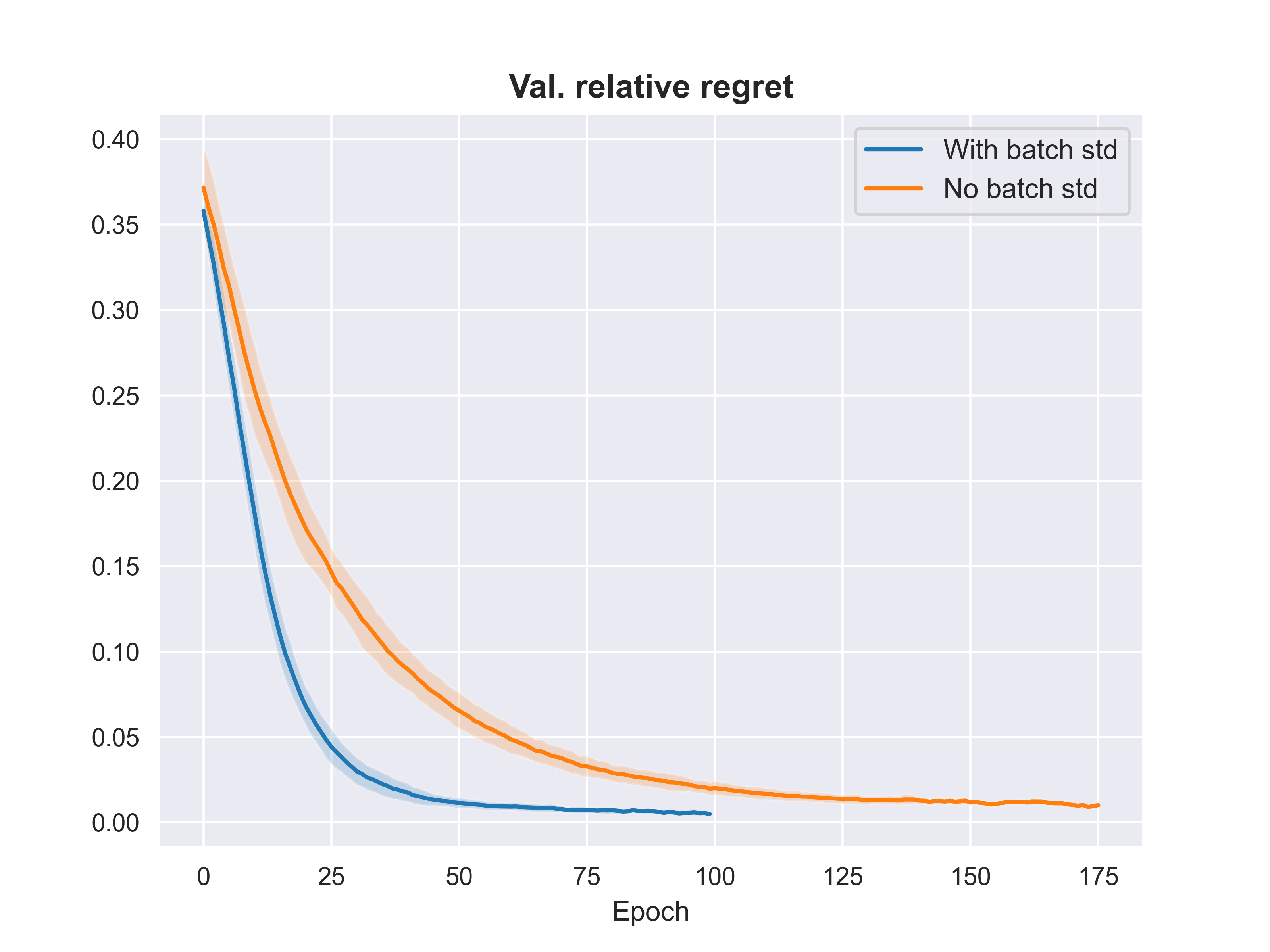}
\caption{Validation relative regret during training of \ouracronym{} with and without standardization on the KP-50}
\label{fig:std_vs_no_std_regret}
\Description{Two learning curves. One of them is associated with SFGE with batch standardization. The other one is associated with SFGE without batch standardization.}
\end{figure}
As previously mentioned in the main body of the paper, we apply standardization to the regret within a single mini-batch to enhance convergence speed by reducing variance in the gradient estimation. The standardization is computed as follows:
\begin{align*}
    \tilde{R} = \frac{R - \mu}{\sigma^2 + \epsilon}
\end{align*}
Here, $R$ represents the (post-hoc) regret, $\mu$ and $\sigma$ denote the mean and variance of the regret within a mini-batch, and $\epsilon=10^{-8}$ is a small constant introduced to prevent numerical instability.
To empirically demonstrate the effectiveness of this standardization operation, we compare a model trained with SFGE with and without the standardization of the regret within mini-batches. We conducted experiments on the KP-50 dataset, following the same evaluation procedure as described in \cref{sec:exp_res}. In \Cref{fig:std_vs_no_std_regret}, we present a comparison of the validation relative regret between the two approaches, clearly illustrating that standardization significantly improves convergence speed.

\begin{figure}[htb]
\centering
\includegraphics[width=0.8\textwidth]{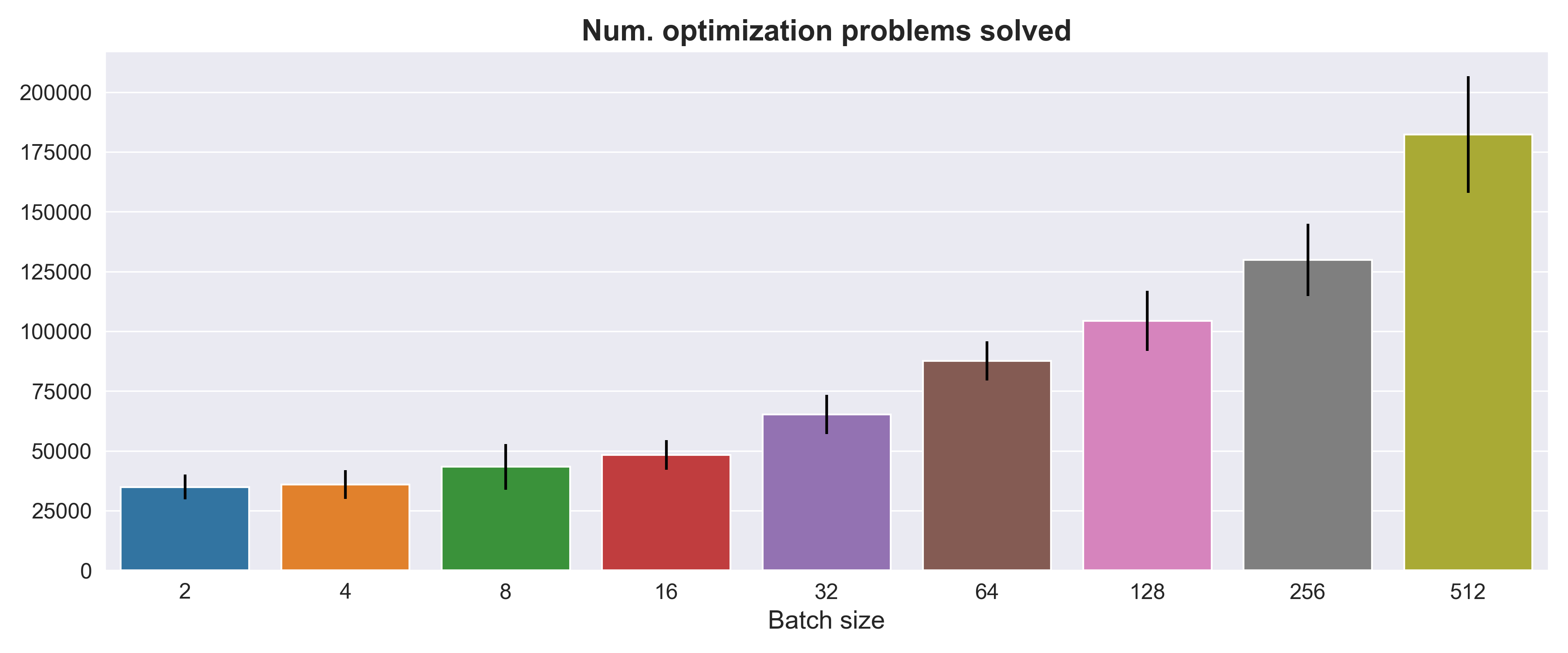}
\caption{Total number of optimization problems solved by \ouracronym{} during training w.r.t. the mini-batch size.}
\label{fig:batch_size_num_probs}
\Description{9 barplots. The height represent the number of epochs. Standard deviation across experiments is represented as a vertical bar.}
\end{figure}
\begin{figure}[htb]
\centering
\includegraphics[width=0.8\textwidth]{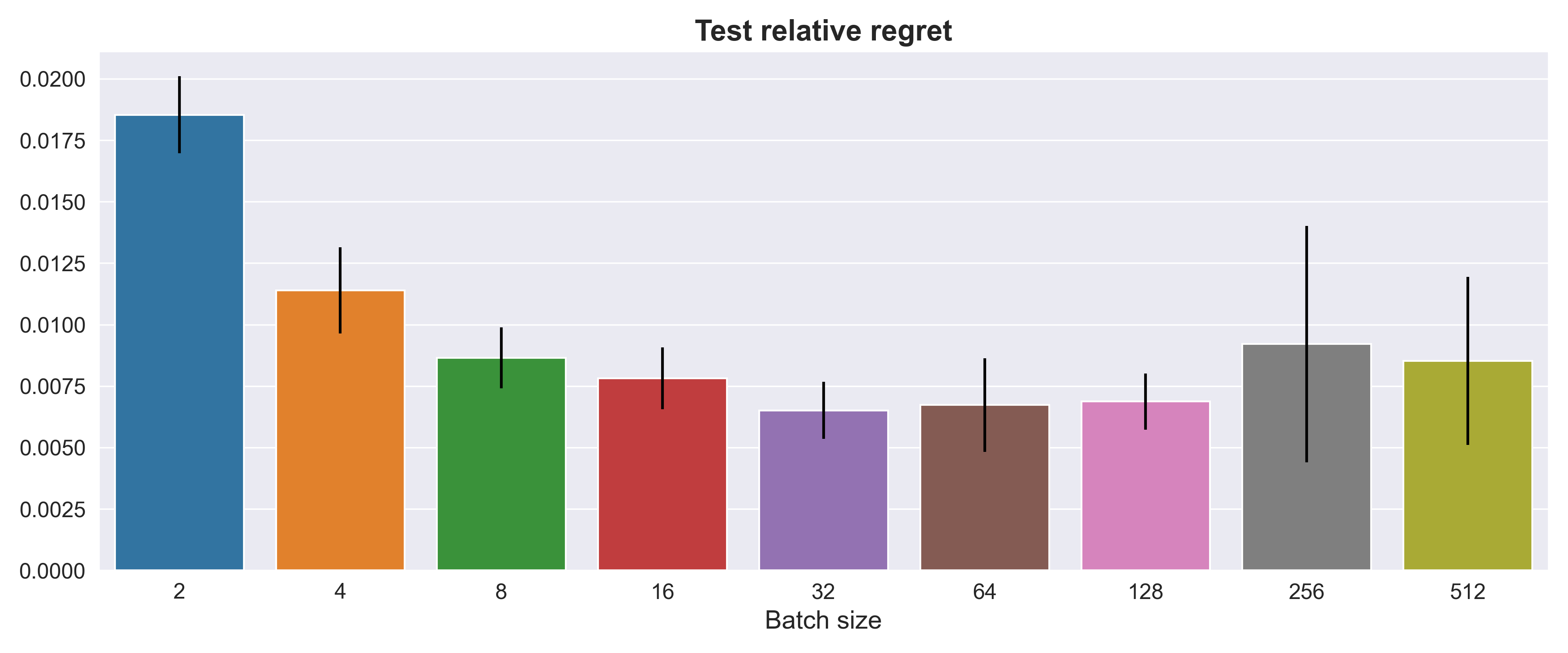}
\caption{\ouracronym{} test relative regret w.r.t. the mini-batch size.}
\label{fig:batch_size_rel_regret}
\Description{9 barplots. The height represents the test relative regret. Standard deviation across experiments is represented with vertical bars.}
\end{figure}

Since the choice of mini-batch size affects the results of standardization and, consequently, the variance reduction, we conducted experiments with various batch sizes, specifically $\{ 2, 4, 8, 16, 32, 64, 128, 256, 512 \}$, on the KP-50 dataset. We evaluated both the relative regret on the test set and the number of optimization problems solved before reaching convergence. The latter experiment provides insights into the computational efficiency of different configurations.
As depicted in \Cref{fig:batch_size_num_probs}, increasing the batch size results in a larger number of optimization problems required to reach convergence, as it necessitates more epochs. With a larger batch size, the number of mini-batches decreases, reducing the number of optimization steps per epoch. Overall, a batch size of 32 demonstrates the best trade-off in terms of computational cost and relative regret.

\section{Comparison with Local-based Approximators in Stochastic Settings} 
\label{appendix:lodl_counter}

In a stochastic setting, where $y$ values are sampled from an unknown distribution depending on the contextual information $x$, such that $y \sim P(Y \mid x)$, regret itself is not deterministic anymore and its expected value on a given point $x_i$ can be expressed as:

\begin{equation}
    \mathbb{E}[\mathit{R}_{x_i}] = \int p(y, x_i) Regret(m_\omega(x_i), y) dy
\end{equation}

Where $m$ is a predictive model parameterized on $\omega$. This implies that the expected global minima for the regret loss function with respect to predictions may differ from the observed ones. Most importantly, a correct method capable of handling the stochastic scenario would have to converge to the expected minimum, assuming to have enough data.

In this section, we will show by means of a counterexample how methods based on local approximations, mainly represented by the work of \citet{Shah0WPT22}, which will be referred to as \textit{LODL}, fail at giving such a guarantee, in contrast with our approach.

\begin{figure}[htb]
\centering
\includegraphics[width=\textwidth]{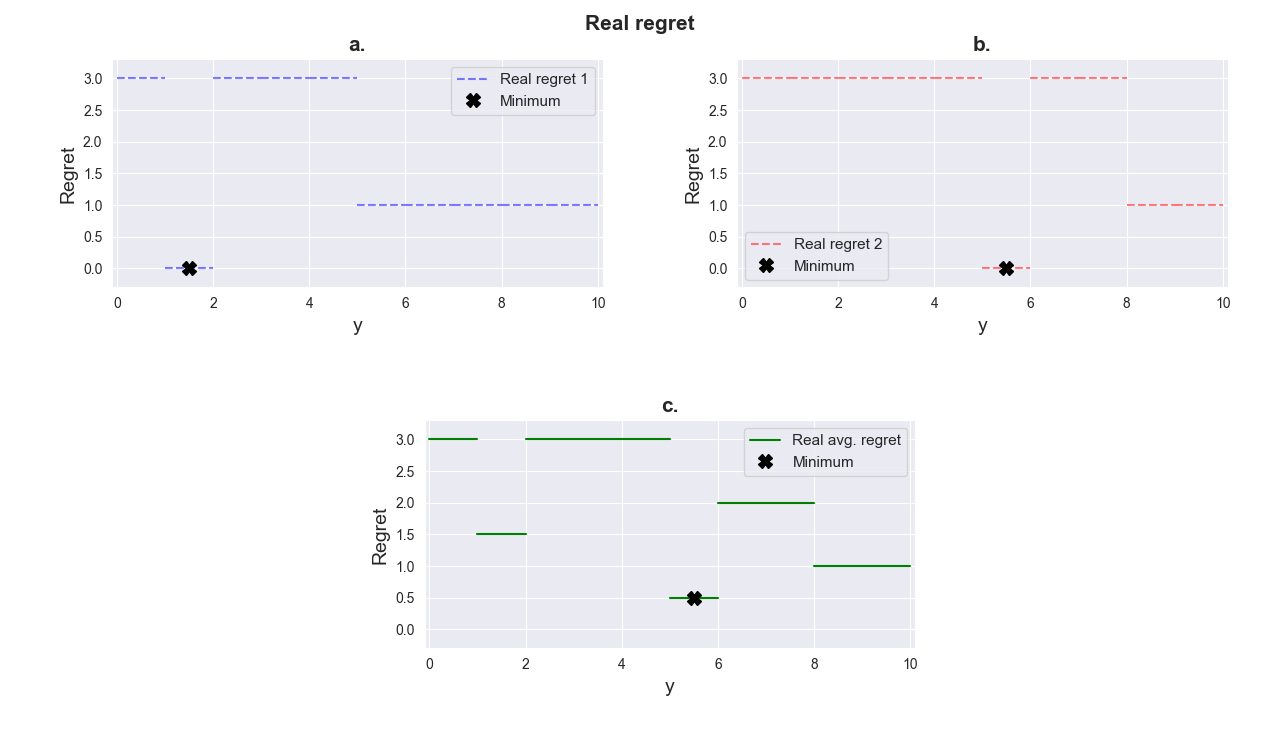}
\caption{Regret functions on a fixed $x_i$ for varying values of $y$. \textbf{a.} Depicts regret $r_1$ for $\hat{y}=y_1$ (first scenario). \textbf{b.} Depicts regret $r_2$ for $\hat{y}=y_2$ (second scenario). \textbf{c.} Depicts the average regret between $r_1$ and $r_2$.}
\label{fig:real_regret_stochastic}
\end{figure}
\begin{figure}[htb]
\centering
\includegraphics[width=0.9\textwidth]{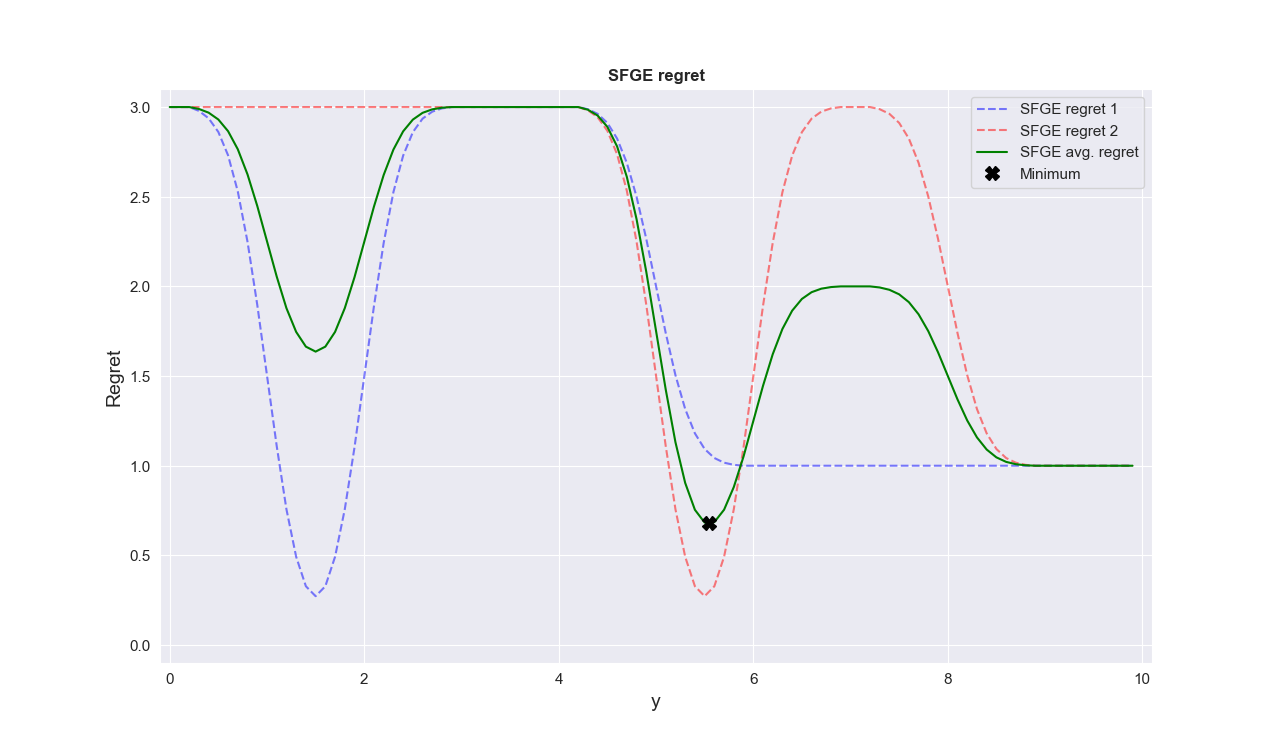}
\caption{Surrogate regret functions obtained by smoothing via SFGE on $r_1$ (blue line) and $r_2$ (red line). The minimum of their average (green line) matches with the expected one.}
\label{fig:sfge_regret_stochastic}
\end{figure}
\begin{figure}[htb]
\centering
\includegraphics[width=0.9\textwidth]{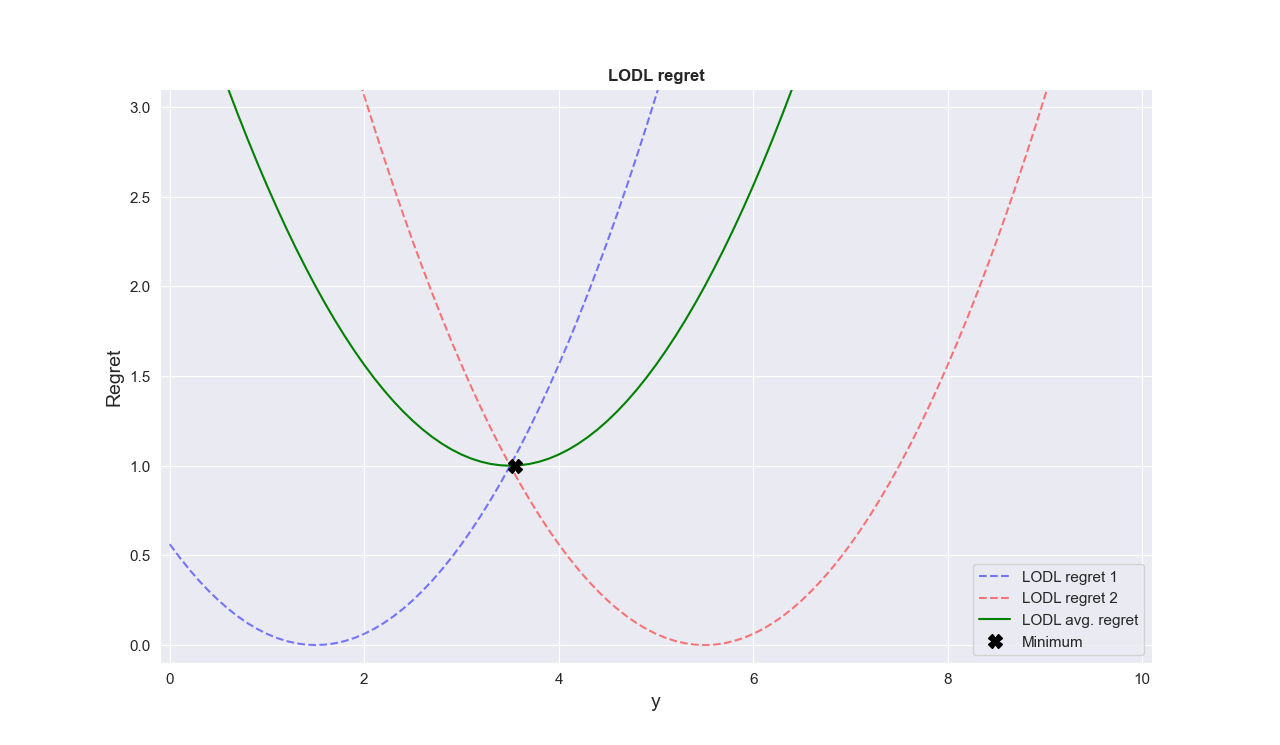}
\caption{Surrogate regret functions obtained by local approximations on $r_1$ (blue line) and $r_2$ (red line). The minimum of their average (green line) is far away from the expected one.}
\label{fig:lodl_regret_stochastic}
\end{figure}

As a simple example, assume to have an instance of a problem, where $x_i$ is fixed and $y \in \mathbb{R}$, with only two possible scenarios:

\begin{enumerate}
    \item $y_1 \in [1.0, 2.0]$, with probability $p(y_1, x_i) = 0.5$
    \item $y_2 \in [5.0, 6.0]$, with equal probability $p(y_2, x_i) = 0.5$
\end{enumerate}

These two possibilities determine two distinct regrets $r_1$ and $r_2$, with one loss landscape being equal to the other, except for a translation along the x-axis, as shown in \Cref{fig:real_regret_stochastic} (\textbf{a.} and \textbf{b.}). Each of them showcases a different global minimum position. The expected regret (\textbf{c.}) is the average of the two (since $y$ is sampled from a uniform distribution) and it is characterized by a different minimum, coinciding with the one of $r_2$, as depicted in \Cref{fig:real_regret_stochastic}.

Given a sufficient number of data points, the expected regret identified by our method is obtained by applying the smoothing procedure to both $r_1$ and $r_2$ and then averaging. Since the smoothing is a global approximation of the original functions, the resulting average is itself a global (smoothed) approximation of the real average regret, guaranteeing the correct location of the optima, up to a certain level of precision, depending on the smoothing strength (e.g., a strong one - with a high value of $\sigma$ - may cause excessive flattening and loss of information). \Cref{fig:sfge_regret_stochastic} shows regret approximations based on this method. The minimum is correctly located at $y=5.5$.

LODL, on the other hand, cannot do the same, as the average of its local approximations does not guarantee to produce a good local approximation of the expected regret around the expected minimum. \Cref{fig:lodl_regret_stochastic} shows how, by fitting two quadratic models around global minima of $r_1$ and $r_2$, their average (another quadratic function, by construction) is still convex, but it misplaces the expected global minimum, which is moved to $y=3.5$.

\clearpage

\section{Behavior of the Learned Standard Deviation During Training}
\label{appendix:sigma_behavior}

In this section, we examine the behavior of the learned standard deviation $\sigma$ of the parameterized Gaussian distribution during training with \ouracronym{}.
For clarity of visualization, we report results on the knapsack problem with 10 and 5 items, where the cost coefficients are unknown.

\begin{figure}[tbh]
\centering
\begin{subfigure}[t]{0.45\textwidth}
  \centering
  \includegraphics[width=\linewidth]{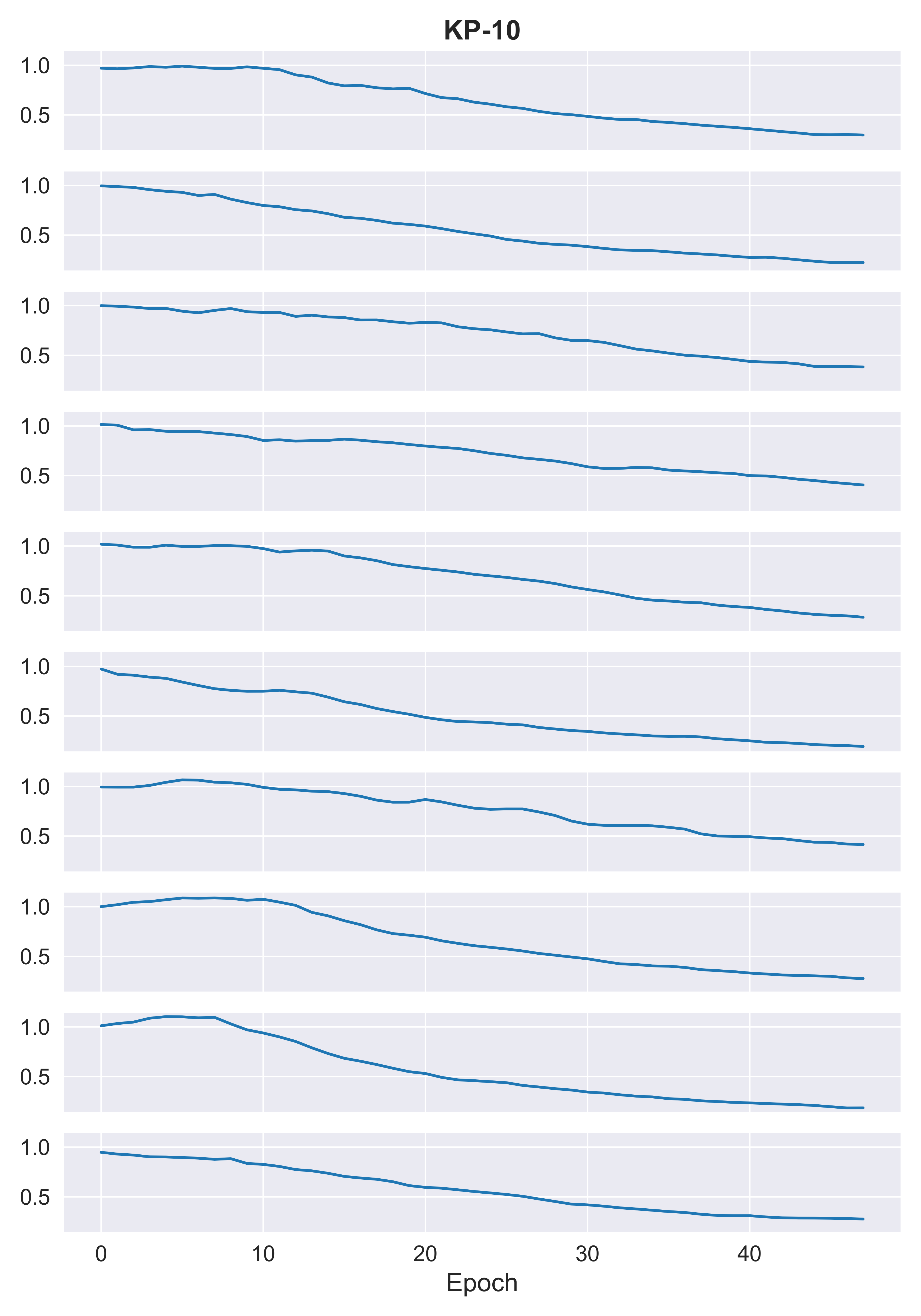}
  \caption{}\label{fig:trainable_std_dev_kp_10}
\end{subfigure}
\hspace{0.5cm}
\begin{subfigure}[t]{0.45\textwidth}
  \centering
  \includegraphics[width=\linewidth]{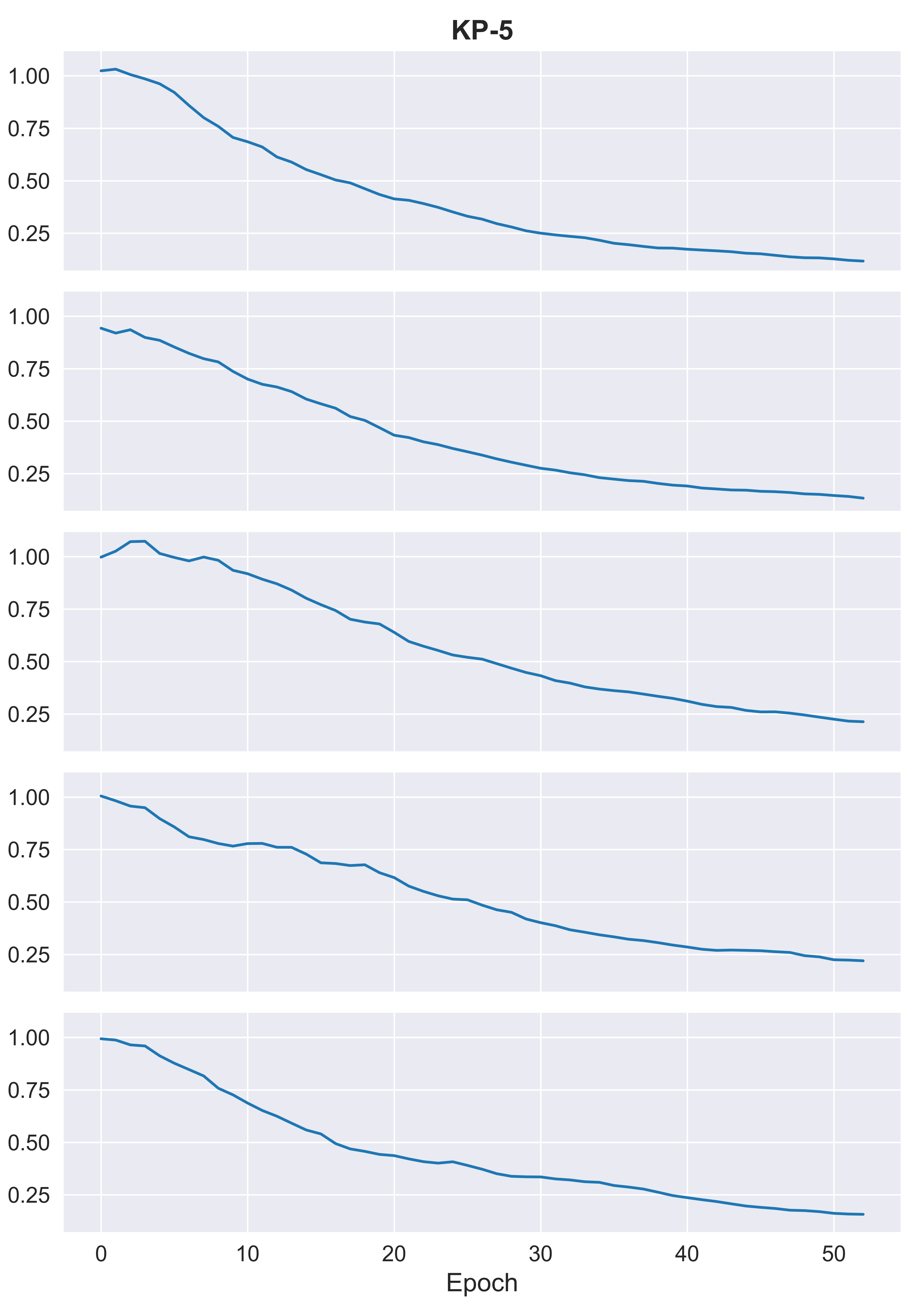}
  \caption{}\label{fig:trainable_std_dev_kp_5}
\end{subfigure}
\caption{The value of the standard deviation of the parametrized Gaussian distribution trained with \ouracronym{} w.r.t. the training epoch on the KP-10 (\subref{fig:trainable_std_dev_kp_10}) and KP-5 (\subref{fig:trainable_std_dev_kp_5}).\label{fig:trainable_std_dev}}
\end{figure}

As shown in \Cref{fig:trainable_std_dev}, $\sigma$ generally decreases over the course of training, though not always monotonically. The decay pattern also varies across cost coefficients. For instance, in KP-10 (\Cref{fig:trainable_std_dev_kp_10}), $\sigma$ gradually decreases for some coefficients, while, for the others, it first exhibits a slight increase during the initial $\approx 10$ epochs before transitioning into a gradual decay.
This might be the result of an initial exploratory phase, which requires a high level of smoothing to obtain more informative gradients.
As training proceeds, $\sigma$ enters a more stable decay, providing a lower level of smoothness and, consequently, the smoothed loss becomes more similar to the original one (the post-hoc regret), thus preserving the true
optimum. In the KP-5, the decay is more pronounced than in KP-10, illustrating that the learning dynamics are problem-dependent. This highlights the advantage of using a trainable $\sigma$: the optimizer automatically adjusts $\sigma$ without requiring a predefined decay schedule. Furthermore, different dimensions of the predicted vector may benefit from distinct decay behaviors, which are naturally captured by this approach.

\section{Relationship among the \ouracronym{} Performance and the Predicted Parameters Size} In this section, we investigate the performance of \ouracronym{} with respect to the predicted parameters size.
We consider the KP problem of sizes 50, 75, 100 and 200, with unknown cost coefficients. The dataset are generated with the same procedure described in \cref{sec:exp_res}. 
We compare PFL, SPO and \ouracronym{}-MAP in terms of relative regret and number of epochs required to reach convergence. 
The hyperparameters are the same described in \cref{sec:exp_res}.

\begin{figure}[htb]
\centering
\includegraphics[width=0.7\textwidth]{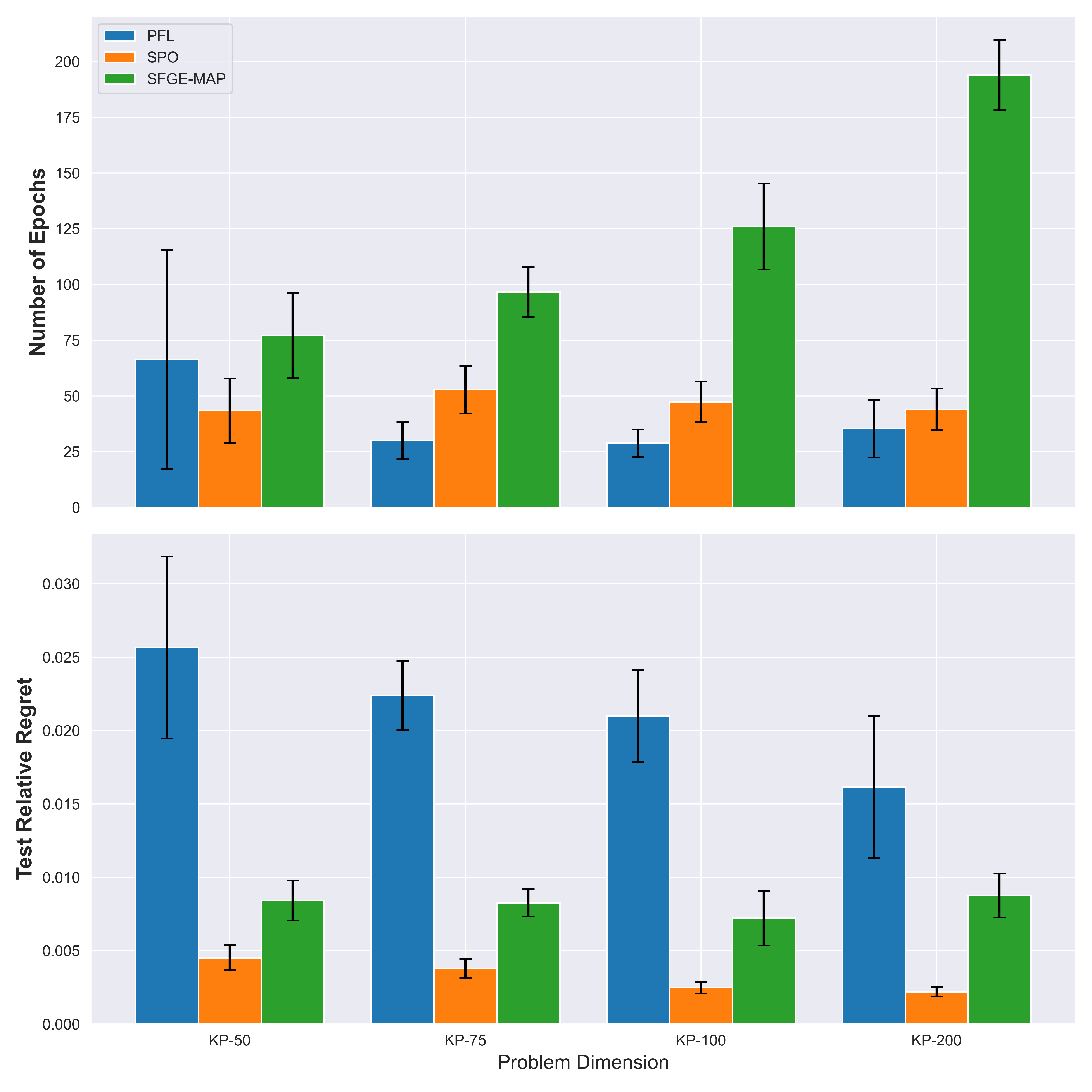}
\caption{Relative regret and number of epochs required to reach convergence w.r.t. the KP problem size.}
\label{fig:performance_wrt_kp_size}
\Description{}
\end{figure}

Results are shown in \Cref{fig:performance_wrt_kp_size}.
While \ouracronym{} maintains stable relative regret across different parameter vector sizes, its convergence speed degrades as dimensionality increases, requiring more epochs for convergence.

\section{Additional Results} 
\label{appendix:supp_res}

In this section, we present supplementary results that further validate and extend the conclusions drawn in the main paper across different problem dimensions and specifications. Specifically, we report results for the following settings: 1) Prediction of the item values for the KP with 75 items and quadratic KP with 8 and 10 items (\Cref{table:knapsack}); 2) Prediction of the item values and weights for the fractional KP with capacity value of 75 (\Cref{appendix_table:frational_kp}); 3) prediction of the capacity value for the KP with unknown capacity value with 50 items (\Cref{table:stochastic_capacity_kp}); 4) prediction of the coverage requirement for the WSMC of size $5 \times 25$ (\Cref{table:wsmc}); 5) prediction of weights for KP with 50 items and stochastic weights, testing on multiple ($30$) samples for each input (\Cref{table:stochastic_weights_kp_multi_samples}): these experiments further demonstrate the ability of \ouracronym{} to converge to a better expected estimation, compared to the other approaches.

\begin{table*}[htb]
\centering
\caption{PFL, \textsc{\ouracronym{}-MAP} and \textsc{SPO} results on the linear and quadratic KP.}
\label{table:knapsack}
\vspace{2pt}
\addtolength{\tabcolsep}{4pt}
\small  
    \begin{tabular}{lccc}
    
    \toprule
    
    \textit{Method} & \textit{Rel. regret} & \textit{MSE ($\times 10^4$)} & \textit{Epochs} \\

    \midrule
    
    \multicolumn{4}{c}{KP-50} \\
    
    \midrule
        
        \textsc{PFL}      & $ 0.022 \pm 0.005 $          & $ \mathbf{2.38  \pm 1.28  } $           & $ \mathbf{40.5 \pm 28.5} $ \\
        \textsc{SPO}      & $ 0.003 \pm 0.0004 $ & $ 4.12 \pm 1.52 $                       & $ \mathbf{40.3 \pm 9.77} $ \\
        \textsc{DPO}      & $ \mathbf{0.002 \pm 0.001} $ & $ 6.24 \pm 3.30 $                       & $ 45.0 \pm 13.4 $ \\
        \ouracronym{}     & $ 0.010 \pm 0.002 $          & $ 17.8  \pm 4.42  $                     & $ 141 \pm 14.0 $ \\
        \ouracronym{}-MAP & $ 0.008 \pm 0.002$           & $ 10.7 \pm 2.03 $                       & $ 83.1 \pm 8.03 $ \\
        \ouracronym{}-MAP (contextual std.dev) & $ 0.008 \pm 0.002$  & $ 9.98 \pm 2.31  $ & $ 89.3 \pm 13.9 $ \\
    
    \midrule

    \multicolumn{4}{c}{KP-75} \\
    
    \midrule
        
        \textsc{PFL}  & $ 0.021 \pm 0.006 $ & $ \mathbf{2.15  \pm 1.11 } $ & $ 44.9 \pm 35.3 $ \\
        \textsc{SPO}  & $ \mathbf{0.003 \pm 0.001} $ & $ 4.24  \pm 1.18 $ & $ 51.5 \pm 10.9 $ \\
        \textsc{DPO}  & $ \mathbf{0.002 \pm 0.0004} $ & $ 5.17  \pm 2.32 $ & $ 45.6 \pm 7.00 $ \\
        \ouracronym{}-MAP & $ 0.008 \pm 0.001 $ & $ 12.6 \pm 4.22  $ & $ 103.8 \pm 14.7 $ \\

    \midrule

    \multicolumn{4}{c}{Quadratic KP-8} \\
    
    \midrule
        
        \textsc{PFL}  & $ 0.034 \pm 0.015 $          & $ \mathbf{2.35  \pm 1.58 } $ & $ \mathbf{24.3 \pm 4.61} $ \\
        \ouracronym{}-MAP & $ 0.006 \pm 0.003 $          & $ 10.6 \pm 6.45  $          & $ 54.5 \pm 14.9 $ \\
        \textsc{SPO}  & $ \mathbf{0.005 \pm 0.002} $ & $ 6.95  \pm 4.09  $          & $ 29.9 \pm 10.4 $ \\

    \midrule

    \multicolumn{4}{c}{Quadratic KP-10} \\
    
    \midrule
        
        \textsc{PFL}  & $ 0.041 \pm 0.011 $          & $ \mathbf{2.37  \pm 1.50 } $ & $ 45.8 \pm 64.0 $ \\
        \ouracronym{}-MAP & $ 0.008 \pm 0.002 $          & $ 8.46  \pm 4.04  $          & $ 54.1 \pm 13.1 $ \\
        \textsc{SPO}  & $ \mathbf{0.006 \pm 0.002} $ & $ 6.47  \pm 3.26  $          & $ 30.7 \pm 8.6 $ \\

    \bottomrule
    
    \end{tabular}
\end{table*} 
\begin{table*}[htb]
\centering
\caption{PFL, \textsc{\ouracronym{}} and \textsc{PO} results on the fractional KP of different sizes and for different penalty coefficient values.}
\label{appendix_table:frational_kp}
\vspace{2pt}
\addtolength{\tabcolsep}{2pt}
\small  
    \begin{tabular}{lccccc}
    
    \toprule
    
    \textit{Method}    & \textit{Rel. PRegret}        & \textit{Feas. rel. PRegret}          & \textit{Infeas. ratio}       & \textit{MSE}                                                    & \textit{Epochs} \\

    \midrule

    \multicolumn{6}{c}{capacity=75, $\rho=0$} \\
    
    \midrule
        
        \textsc{PFL}    & $ 0.353 \pm 0.014 $          & $ \mathbf{0.096 \pm 0.058} $                & $ \mathbf{0.72 \pm 0.15} $     & $ \mathbf{99.1 \pm 13.1} $   & $13.3 \pm 2.9$ \\
        \textsc{SFGE}   & $ 0.337 \pm 0.009 $          & $ - $       & $ 0.99 \pm 0.01 $             & $ 6.20 \cdot 10^5 \pm 6.06 \cdot 10^5 $                     & $ 13.4 \pm 4.1 $ \\
        \textsc{P+O}    & $0.332 \pm 0.109 $  & $ - $                              & $ 1.0 \pm 0.0 $               & $ 9.8 \cdot 10^5 \pm 1.1 \cdot 10^4 $                      & $ \mathbf{2.4 \pm 1.4} $ \\

    \midrule

    \multicolumn{6}{c}{capacity=75, $\rho=1$} \\
    
    \midrule
        
        \textsc{PFL}    & $ 0.437 \pm 0.023 $          & $ \mathbf{0.096 \pm 0.058} $                & $ 0.72 \pm 0.15 $               & $ \mathbf{99.1 \pm 13.1} $   & $13.3 \pm 2.87$ \\    
        \textsc{SFGE}   & $ 0.410 \pm 0.010 $          & $ 0.172 \pm 0.044 $                         & $ \mathbf{0.52 \pm 0.09} $     & $ 9.41 \cdot 10^5 \pm 5.40 \cdot 10^5 $                                    & $ 16.3 \pm 3.8 $ \\
        
        \textsc{P+O}    & $0.405 \pm 0.145 $  & $ 0.332 \pm 0.013 $                         & $ 0.617 \pm 0.035 $          & $ 3.8 \cdot 10^5 \pm 4.5 \cdot 10^3 $                                     & $ \mathbf{1.8 \pm 1.5} $ \\
    
    \midrule
    
    \multicolumn{6}{c}{capacity=75, $\rho=2$} \\
    
    \midrule
        
        \textsc{PFL}    & $ 0.522 \pm 0.057 $           & $ \mathbf{0.096 \pm 0.058} $                & $ 0.72 \pm 0.15 $     & $ \mathbf{99.1 \pm 13.1} $    & $13.3 \pm 2.87$ \\   
        \textsc{SFGE}   & $ 0.436 \pm 0.010 $           & $ 0.230 \pm 0.081 $                         & $ \mathbf{0.40 \pm 0.16} $     & $ 2.1 \cdot 10^6 \pm 1.53 \cdot 10^6 $                                      & $ 20.8 \pm 7.8 $ \\
        \textsc{P+O}    & $0.426 \pm 0.149$   & $ 0.378 \pm 0.009 $                         & $ 0.428 \pm 0.025 $         & $ 3.5 \cdot 10^5 \pm 4.0 \cdot 10^3 $                                       & $ \mathbf{3.2 \pm 2.0} $ \\

    \bottomrule
    
    \end{tabular}
\end{table*} 
\begin{table*}[tb]
\centering
\vspace*{1.5\baselineskip}

\caption{\textsc{PFL} and \ouracronym{} results on the KP with uncertain capacity of different sizes and for different penalty coefficient values.}\label{table:stochastic_capacity_kp}
    \vspace{2pt}
    \addtolength{\tabcolsep}{2pt}
    \small
    \begin{tabular}{lcccc}
    \toprule
    \textit{Method} & \textit{Rel. PRegret} & \textit{Infeas. ratio} & \textit{MSE} & \textit{Epochs} \\
    \midrule
    
        \multicolumn{5}{c}{50-items, $\rho=5$} \\
        \midrule
        PFL & 0.556 $\pm$ 0.353 & \textbf{0.667 $\pm$ 0.257} & \textbf{10.81 $\pm$ 6.30} & \textbf{11.5 $\pm$ 1.9} \\
        \midrule
        SFGE & \textbf{0.367 $\pm$ 0.238} & 0.752 $\pm$ 0.298 & 16.82 $\pm$ 12.00 & 44.9 $\pm$ 14.5 \\
        \midrule
        
        \multicolumn{5}{c}{50-items, $\rho=10$} \\
        \midrule
        PFL & 1.213 $\pm$ 0.780 & \textbf{0.667 $\pm$ 0.257} & \textbf{10.81 $\pm$ 6.30} & \textbf{11.5 $\pm$ 1.9} \\
        \midrule
        SFGE & \textbf{0.556 $\pm$ 0.322} & 0.893 $\pm$ 0.076 & 23.24 $\pm$ 16.78 & 57.2 $\pm$ 30.4 \\
        \midrule
        
        \multicolumn{5}{c}{50-items, $\rho=20$} \\
        \midrule
        PFL & 2.520 $\pm$ 1.631 & \textbf{0.667 $\pm$ 0.257} & \textbf{10.812 $\pm$ 6.295} & \textbf{11.5 $\pm$ 1.9} \\
        \midrule
        SFGE & \textbf{0.800 $\pm$ 0.466} & 0.953 $\pm$ 0.031 & 32.415 $\pm$ 22.131 & 48.1 $\pm$ 21.1 \\
        
    \bottomrule
    \end{tabular}
    \end{table*}
\begin{table*}[htb]
\centering
\caption{PFL and \ouracronym{} results on the WSMC.}
\label{table:wsmc}
\vspace{2pt}
\addtolength{\tabcolsep}{4pt}
\small
    \begin{tabular}{lccccc}
    
    \toprule

    \textit{Method}  &  \textit{Rel. PRegret}  & \textit{Feas. rel. PRegret}  &  \textit{Infeas. ratio}  &  \textit{MSE}         &  \textit{Epochs} \\
    
    \midrule
    
    \multicolumn{6}{c}{$5 \times 25$, $\rho=1$} \\
    
    \midrule

        \textsc{PFL}              & $1.18 \pm 0.62 $            & $ \mathbf{0.154 \pm 0.076} $          & $ 0.63 \pm 0.11 $            & $ \mathbf{4.74 \cdot 10^4 \pm 2.10 \cdot 10^4} $             & $ \mathbf{42.8 \pm 23.1} $ \\
        \textsc{CombOptNet}       & $ 4.05 \pm 2.58 $           & $ - $                                 & $ 0.97 \pm 0.04 $            & $ 6.20 \cdot 10^6 \pm 5.24 \cdot 10^6 $                           & $ 57.4 \pm 30.8  $ \\
        \textsc{SFGE}             & $\mathbf{0.850 \pm 0.264} $ & $ 0.314 \pm 0.361 $                   & $ \mathbf{0.55 \pm 0.27} $   & $ 1.80 \cdot 10^5 \pm 6.85 \cdot 10^4 $                           & $ 55.2 \pm 15.5 $ \\
    
    \midrule
    
    \multicolumn{6}{c}{$5 \times 25$, $\rho=5$} \\
    
    \midrule
        
        \textsc{PFL}             & $7.27 \pm 4.20$             & $\mathbf{0.182 \pm 0.037}$             & $ 0.60 \pm 0.11 $            & $ \mathbf{8.03 \cdot 10^4 \pm 2.08 \cdot 10^4} $            & $ 49.5 \pm 24.2 $ \\
        \textsc{CombOptNet}      & $ 149.76 \pm 91.56 $        & $ - $                                  & $ 0.99 \pm 0.02 $            & $ 9.08 \cdot 10^6 \pm 3.87 \cdot 10^6 $                          & $ \mathbf{40.1 \pm 19.1} $ \\
        \textsc{SFGE}            & $\mathbf{2.53 \pm 0.53}$    & $1.28 \pm 0.44$                        & $\mathbf{0.23 \pm 0.11}$     & $ 4.70 \cdot 10^5 \pm 1.94 \cdot 10^5 $ & $48.8 \pm 10.2$ \\

    \midrule
    
    \multicolumn{6}{c}{$5 \times 25$, $\rho=10$} \\

    \midrule
        
        \textsc{PFL}  & $16.0 \pm 10.1$           & $\mathbf{0.12 \pm 0.03}$                 & $ 0.67 \pm 0.07 $             & $ \mathbf{7.55 \cdot 10^4 \pm 5.65 \cdot 10^4} $                & $ \mathbf{37.9 \pm 16.3} $ \\
        \textsc{CombOptNet}                       & $ 602.1 \pm 276.7 $                    & $ - $                         & $ 0.98 \pm 0.06 $            
                      & $ 1.27 \cdot 10^7 \pm 1.55 \cdot 10^7 $                              & $ 47.6 \pm 26.5  $ \\
        \textsc{SFGE} & $\mathbf{3.00 \pm 0.50}$  & $1.72 \pm 0.41$                          & $\mathbf{0.12 \pm 0.07}$     & $ 4.19 \cdot 10^5 \pm 1.06 \cdot 10^5 $                         & $47.5 \pm 12.8$ \\

    \bottomrule
    
    \end{tabular}

\end{table*} 

\begin{table*}[tb]
\centering
\vspace*{1.5\baselineskip}

    \caption{\textsc{PFL} and \ouracronym{} results on the KP-50 with uncertain weights and $30$ $y$ samples for each $x$. We omit the \textit{Feas. rel. PRegret} for \textit{Infeas. ratios} near 1.}
    \label{table:stochastic_weights_kp_multi_samples}
    \vskip 0.1in
    \addtolength{\tabcolsep}{2pt}
    \small
    \begin{tabular}{lccccc}
    \toprule
    \textit{Method} & \textit{Rel. PRegret} & \textit{Feas. rel. PRegret} & \textit{Infeas. ratio} & \textit{MSE} & \textit{Epochs} \\
    \midrule
    
        \multicolumn{6}{c}{50-items, $\rho=5$} \\
        \midrule
        PFL                 & $0.614 \pm 0.128$           & $0.001 \pm 0.001$ & $0.93 \pm 0.03$ & $\mathbf{1.73 \cdot 10^5 \pm 3.23 \cdot 10^4}$ & $\mathbf{34.0 \pm 20.3}$ \\
        \textsc{CombOptNet} & $0.719 \pm 0.244$           & $0.008 \pm 0.005$ & $0.91 \pm 0.02$ & $2.67 \cdot 10^8 \pm 8.02 \cdot 10^7$          & $41.0 \pm 13.4$ \\
        SFGE(ours)          & $\mathbf{0.374 \pm 0.061}$  & - & $0.94 \pm 0.02$ & $2.88 \cdot 10^5 \pm 8.79 \cdot 10^4$           & $136.0 \pm 10.8$ \\
        \midrule

        \multicolumn{6}{c}{50-items, $\rho=10$} \\
        \midrule
        PFL                 & $1.218 \pm 0.242$           & $0.001 \pm 0.001$ & $0.93 \pm 0.03$ & $\mathbf{1.73 \cdot 10^5 \pm 3.23 \cdot 10^4}$ & $\mathbf{34.0 \pm 20.3}$ \\
        \textsc{CombOptNet} & $1.685 \pm 0.501$           & $0.008 \pm 0.005$ & $0.91 \pm 0.02$ & $2.67 \cdot 10^8 \pm 8.02 \cdot 10^7$          & $41.0 \pm 13.4$ \\
        SFGE(ours)          & $\mathbf{0.736 \pm 0.103}$  & - & $0.94 \pm 0.02$ & $3.06 \cdot 10^5 \pm 9.23 \cdot 10^4$           & $128.8 \pm 19.2$ \\
        \midrule

        \multicolumn{6}{c}{50-items, $\rho=20$} \\
        \midrule
        PFL                 & $2.418 \pm 0.481$           & $0.001 \pm 0.001$ & $0.93 \pm 0.03$ & $\mathbf{1.73 \cdot 10^5 \pm 3.23 \cdot 10^4}$ & $\mathbf{34.0 \pm 20.3}$ \\
        \textsc{CombOptNet} & $3.178 \pm 0.989$           & $0.008 \pm 0.005$ & $0.91 \pm 0.02$ & $2.67 \cdot 10^8 \pm 8.02 \cdot 10^7$          & $41.0 \pm 13.4$ \\
        SFGE(ours)          & $\mathbf{1.106 \pm 0.164}$  & - & $0.94 \pm 0.02$ & $3.12 \cdot 10^5 \pm 9.52 \cdot 10^4$           & $119.3 \pm 26.1$ \\

    \bottomrule
    \end{tabular}
\end{table*}
\begin{figure}[htb]
    \centering\includegraphics[width=\textwidth]{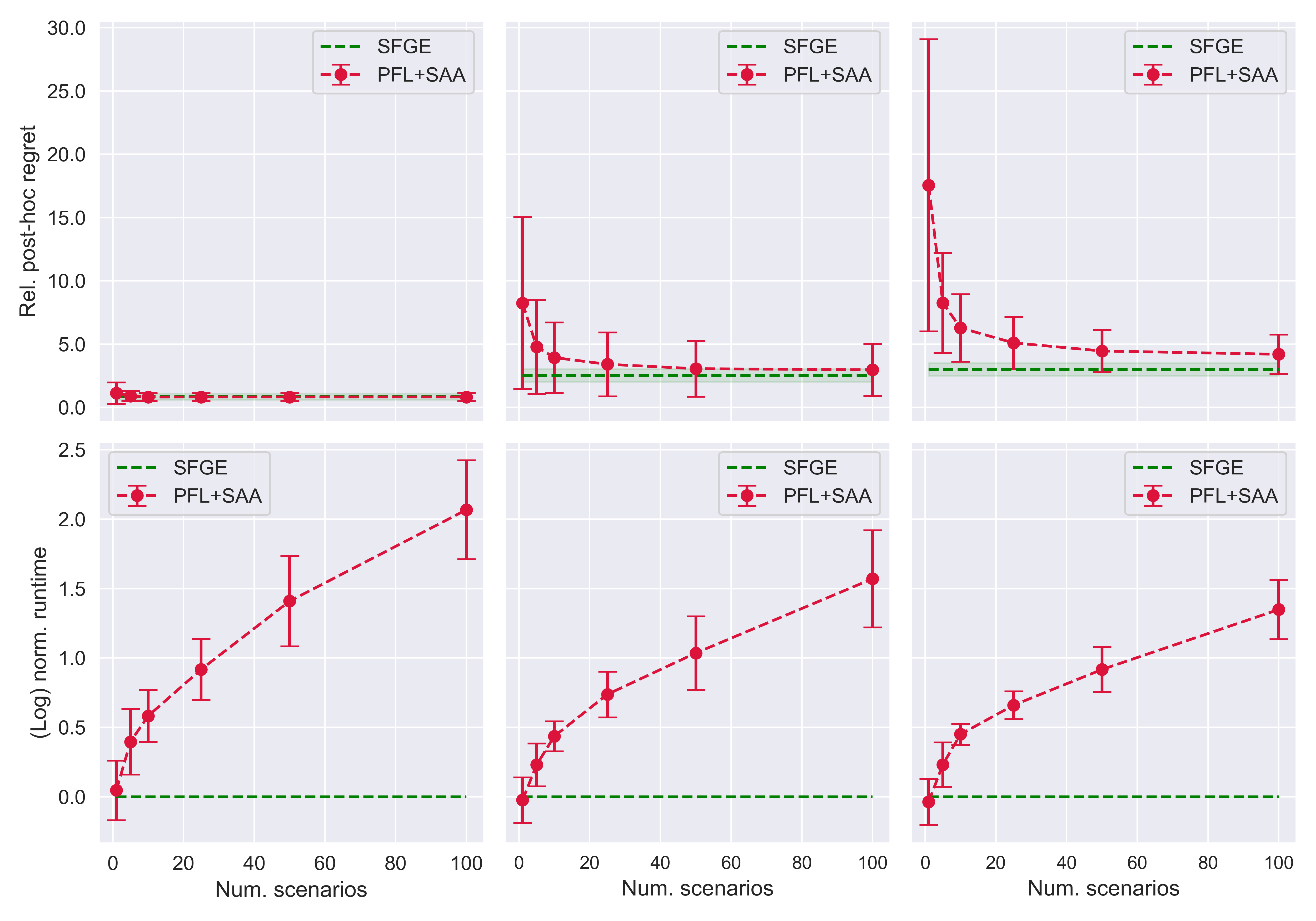}
    \caption{Comparison between \ouracronym{} and PFL+SAA on the WSMC of size $5 \times 25$ for $\rho=1$ (left), $\rho=5$ (center) and $\rho=10$ (right).}
    \label{fig:wsmc_5_25}
\end{figure}
\clearpage
\begin{figure}[th]
    \centering
    \includegraphics[width=\textwidth]{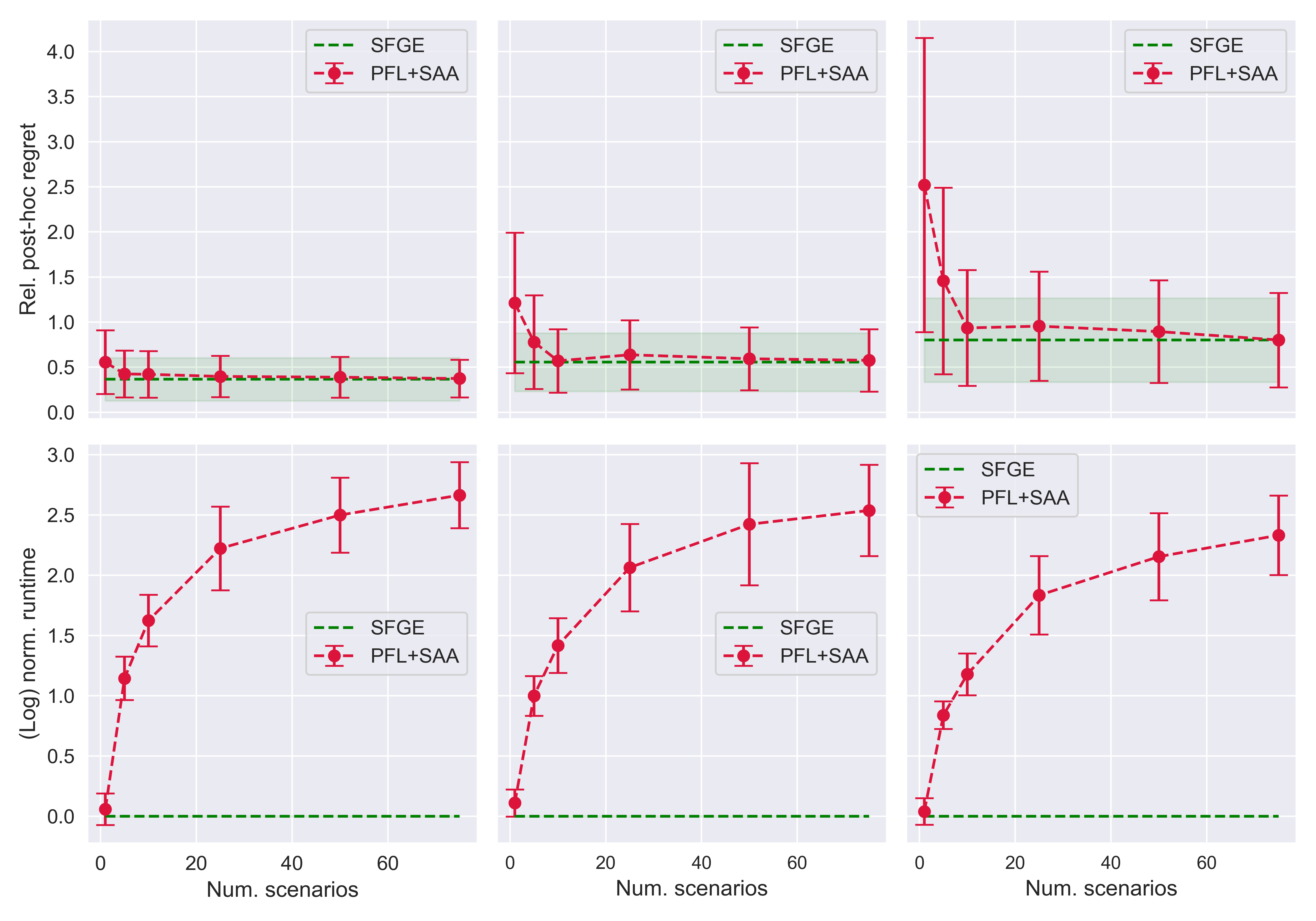}
    \caption{Comparison between \ouracronym{} and PFL+SAA on the KP-50 with stochastic capacity, for $\rho=5$ (left), $\rho=10$ (center) and $\rho=20$ (right).}
    \label{fig:stochastic_capacity_kp_50_items}
\end{figure}
%



\end{document}